\documentclass[11pt,a4paper]{article}
\usepackage{jinstpub} 
\usepackage{lineno}
\usepackage{tabularx}
\usepackage{booktabs}
\usepackage{multirow}
\usepackage{microtype}
\usepackage{soul}
\usepackage{subcaption}
\usepackage{hyphenat}
\usepackage{xurl}
\usepackage{orcidlink}

\proceeding{4$^{\text{th}}$ Artificial Intelligence for the Electron Ion Collider (2025)\\
27 October 2025--29 October 2025\\
MIT, Boston, USA}

\title{ScatterPrism: convergence for generative simulation and inverse problems in particle and nuclear physics%
\note[$\ast$]{This is the Author Accepted Manuscript. The Version of Record is published in \textit{Journal of Instrumentation} \textbf{21} (2026) C07012, doi:\href{https://doi.org/10.1088/1748-0221/21/07/C07012}{10.1088/1748-0221/21/07/C07012}, and is distributed under the terms of the Creative Commons Attribution 4.0 International License (\href{https://creativecommons.org/licenses/by/4.0/}{CC BY 4.0}).}}

\author[a,1]{Zeyu~Xia\,\orcidlink{0000-0003-0234-5857},%
\note{Corresponding author.}}
\author[a]{Tyler~Kim\,\orcidlink{0009-0002-2989-9745},}
\author[b]{Trevor~Reed\,\orcidlink{0000-0002-4230-6687},}
\author[b]{Judy~Fox\,\orcidlink{0000-0001-8198-4117},}
\author[a]{Geoffrey~Fox\,\orcidlink{0000-0003-1017-1391}}
\author[c]{and Adam~Szczepaniak\,\orcidlink{0000-0002-4156-5492}}

\affiliation[a]{Department of Computer Science, University of Virginia,\\
85 Engineer's Way, Charlottesville, VA 22904, USA}

\affiliation[b]{School of Data Science, University of Virginia,\\
1919 Ivy Road, Charlottesville, VA 22903, USA}

\affiliation[c]{Department of Physics, Indiana University Bloomington,\\
727 East Third Street, Bloomington, IN 47405, USA}

\emailAdd{zeyu.xia@virginia.edu}

\abstract{
High-fidelity simulations and complex inverse problems, such as detector modeling and unfolding, are computationally intensive bottlenecks across subatomic physics, yet essential for accurate physical interpretation. While Conditional Flow Matching (CFM) offers a robust acceleration approach, we demonstrate its standard training loss is fundamentally misleading. Specifically, utilizing a Jefferson Lab Nuclear Physics (NP) kinematic dataset ($\gamma p \to \rho^0 p \to \pi^+\pi^- p$), we expose that CFM loss plateaus prematurely, obscuring ongoing physical refinement. To verify this disconnect is a dataset-agnostic pathology, we introduce ScatterPrism, an efficient generative surrogate evaluated against both the NP data and synthetic stress tests modeling challenging 1D distribution topologies. Coupling these benchmarks, we establish that physics-informed metrics continue improving long after standard loss converges. Consequently, we propose a multi-metric diagnostic protocol to ensure true kinematic fidelity without data memorization. Driven by NP challenges relevant to the forthcoming Electron-Ion Collider (EIC), this unified machinery has strong potential to extend to High-Energy Physics (HEP) applications, such as jet modeling. Furthermore, the framework holds promise for broader domains requiring rigorous generative reliability, including medical imaging, astrophysics, and quantitative finance.
}


\arxivnumber{2604.01313} 

\begin{document}
\maketitle
\flushbottom

\section{Introduction}%
\label{sec:intro}

Modern experimental physics analyses rely on large-scale Monte Carlo simulation datasets to attain the statistical precision required for high-fidelity measurements. While parton-level event generation (e.g., Pythia~\cite{bierlichComprehensiveGuidePhysics2022}, MadGraph~\cite{alwallMadGraph5Going2011}) is comparatively cheap per event, the subsequent full detector-response simulation (typically GEANT4~\cite{agostinelliGeant4aSimulationToolkit2003}) dominates compute and scales unfavorably with event volume and detector complexity. This challenge is rooted in contemporary Nuclear Physics (NP) programs, such as Jefferson Lab experiments (e.g., CLAS12~\cite{burkertCLAS12SpectrometerJefferson2020}, GlueX~\cite{adhikariGlueXBeamlineDetector2021}) and the forthcoming Electron-Ion Collider (EIC), where modeling particle transport and detector responses dominates computational budgets. These barriers are shared by High-Energy Physics (HEP) initiatives like the Large Hadron Collider (LHC), establishing a universal need for faster AI-based surrogate models across both communities.

Detector unfolding---recovering true event-level observables from detector-level measurements degraded by detector effects, such as finite instrumental resolution---further magnifies these computational demands, motivating the shift toward generative deep learning. Deep generative models, particularly Conditional Flow Matching (CFM), offer stable, simulation-free training. However, using the JLab NP photoproduction dataset, we demonstrate that standard CFM training loss is an unreliable indicator of true physical convergence.

This limitation, characterized as spectral bias~\cite{wangAnalyticalTheorySpectral2026}, obscures ongoing physical refinement. To resolve this dataset-agnostic pathology, we introduce ScatterPrism, a CFM framework tailored for high-fidelity kinematic event generation and detector unfolding. Our primary contributions are:

\begin{enumerate}
    \item \emph{Convergence diagnostics and validation suite.} We identify the premature plateau of standard CFM loss, which obscures ongoing physical refinement, and establish a rigorous multi-metric protocol to accurately track true convergence, verify generative fidelity, and prevent data memorization; the constituent metrics are introduced and motivated in Section~\ref{subsec:metrics}.

    \item \emph{ScatterPrism Framework.} We introduce ScatterPrism, a configurable CFM-based tool, and validate its capabilities for event generation and conditional detector unfolding on a realistic Jefferson Lab dataset ($\gamma p \to \rho^0 p \to \pi^+\pi^- p$) relevant to the forthcoming EIC.

    \item \emph{Synthetic stress tests.} We provide controlled 1D benchmarks (\texttt{gaussian}, \texttt{high-cut}, \texttt{multi-peak}, \texttt{high-frequency}, \texttt{delta}, \texttt{uniform}, and \texttt{exponential}) to isolate generative capabilities and diagnose topological failure modes prior to deployment on real physics data.
\end{enumerate}

\section{Related work}%
\label{sec:related}

Two computational pillars dominate subatomic-physics data pipelines: forward generative simulation~\cite{hashemiDeepGenerativeModels2024} and detector unfolding~\cite{huetschLandscapeUnfoldingMachine2025}. Both are bottlenecked by computationally expensive GEANT-class algorithms, motivating parallel machine-learning surrogates. On the forward simulation side, architectures have evolved from Generative Adversarial Network (GAN)-based shower simulators (CaloGAN~\cite{paganiniCaloGANSimulating3D2018}) and phase-space samplers~\cite{butterHowGANLHC2019} to high-fidelity normalizing flows like CaloFlow~\cite{krauseFastAccurateSimulations2023}. Recent benchmarks systematically validate these fast-simulation surrogates against full generation engines like GEANT4~\cite{ahmadComprehensiveEvaluationGenerative2024}.

On the inverse side, traditional binned unfolding methods like Iterative Bayesian Unfolding (IBU) suffer from dimensionality curses. OmniFold~\cite{andreassenOmniFoldMethodSimultaneously2020} circumvented this via unbinned neural classifiers, initiating a shift toward deep generative unfolding. Modern approaches include conditional invertible networks (cINNs)~\cite{backesUnfoldingMethodBased2024} and Schr\"{o}dinger-bridge formulations~\cite{diefenbacherImprovingGenerativeModelbased2024}. These domains increasingly converge on a shared toolkit: while diffusion models~\cite{hoDenoisingDiffusionProbabilistic2020} define transport through stochastic transitions and scale well using denoising or score-matching objectives, Conditional Flow Matching~\cite{lipmanFlowMatchingGenerative2022} directly regresses the transport vector field that generates a probability path under a deterministic flow. Building on this, ScatterPrism differentiates itself from likelihood-based cINNs~\cite{backesUnfoldingMethodBased2024} by learning a CFM velocity field end-to-end without iterative refinement.

These foundations rapidly populate workflows in both fields, with modern generative architectures yielding analysis-ready unfolding for complex final states~\cite{shmakovFullEventParticlelevel2025}. While generative Artificial Intelligence (AI) is vital for EIC simulations~\cite{allaireArtificialIntelligenceElectron2024}, current implementations often assume generic metrics safely indicate modeling quality. A crucial gap remains: no prior work systematically addresses the disconnect between training loss convergence and true kinematic fidelity of CFM models, a concern relevant to both NP and HEP. We address this gap on the NP side using a low-multiplicity JLab photoproduction dataset; the diagnostic methodology is designed to transfer to HEP settings, though further validation is left to future work.

\section{Methodology}%
\label{sec:method}

\subsection{Datasets and feature representation}%
\label{subsec:datasets}

\paragraph{MC-POM dataset.}

The MC-POM (Monte Carlo Pomeron) dataset models exclusive photoproduction $\gamma p \rightarrow \rho^0 p \rightarrow \pi^+\pi^- p$. This serves as a representative low-multiplicity NP topology. We focus on forward kinematics with low momentum transfer ($|t| < 1~\mathrm{GeV}^2$), where pomeron exchange dominates. In this regime, the P-wave $\rho(770)$ resonance is prominent in the $M(\pi^+\pi^-)$ mass spectrum. We utilize a fixed dataset of 8M events, partitioned into an 8:1:1 training/validation/test split (6.4M / 0.8M / 0.8M events) used consistently across all generation and unfolding experiments.

Events initially consist of 24-dimensional vectors encoding the four-momenta of all involved particles ($p_\gamma^\mu, p_1^\mu, p_2^\mu, \pi^\pm$) and derived variables ($t, M_{\pi\pi}, \cos\theta, \phi$) evaluated in the $\pi^+\pi^-$ helicity rest frame. To eliminate redundancy, we project the data into a 10-dimensional phase space. Excluding the recoil proton (fixed by four-momentum conservation) and extracting the spatial momenta $(p_x, p_y, p_z)$ of the remaining four particles yields 12 components. Dropping the identically zero $p_y$ components of the incident photon and target proton results in the final 10 dimensions. Each feature was standardized to zero mean and unit variance, then scaled by 5.0; this empirically outperformed factors of 1.0 and 2.0. The mismatch with the unit-variance prior $\mathcal{N}(0, I)$ separates source and target supports along $x_t = t\,x_1 + (1{-}t)\,x_0$, amplifying the deterministic velocity signal and improving training stability. The inverse transform recovers physical units.

For detector unfolding, we simulate resolution effects via independent Gaussian smearing of each Cartesian momentum component $k$ of the $\pi^\pm$ tracks, with standard deviation $k^2 \cdot \sigma_{\mathrm{smear}}$ for $\sigma_{\mathrm{smear}} \in \{0.5, 1.0, 2.0\}$. The recoil proton four-momentum is subsequently algebraically inferred ($p_2^\mu = p_\gamma^\mu + p_1^\mu - p_{\pi^+}^\mu - p_{\pi^-}^\mu$). Because pion smearing directly breaks energy-momentum conservation, this inferred state drops off the exact invariant mass shell. A primary advantage of ScatterPrism is its ability to ingest this physically `broken' conditional data and learn the implicit constraints required to project it back onto the exact ground-truth manifold. Throughout training, all physics-informed metrics are monitored on the held-out validation split, and final results in Section~\ref{subsec:mcpom_results} are reported on the held-out test split. We compute the nearest-neighbor ratio $R_{\mathrm{NN}}$, evaluated against the training manifold, as an explicit guard against memorization.

\paragraph{Synthetic mock datasets.}

To isolate modeling challenges, we construct a diverse suite of 1D synthetic benchmarks (\texttt{gaussian}, \texttt{high-cut}, \texttt{multi-peak}, \texttt{high-frequency}, \texttt{delta}, \texttt{uniform}, and \texttt{exponential}) across configuration presets. These controlled environments enable rigorous ablation studies of mode collapse and fine-grained resolution prior to real physics deployment. Detailed formulations and extended results are in Appendix~\ref{app:synthetic}.

\subsection{Conditional flow matching framework}%
\label{subsec:cfm_framework}

\begin{figure}[htbp]
\centering
\includegraphics[width=0.8\textwidth]{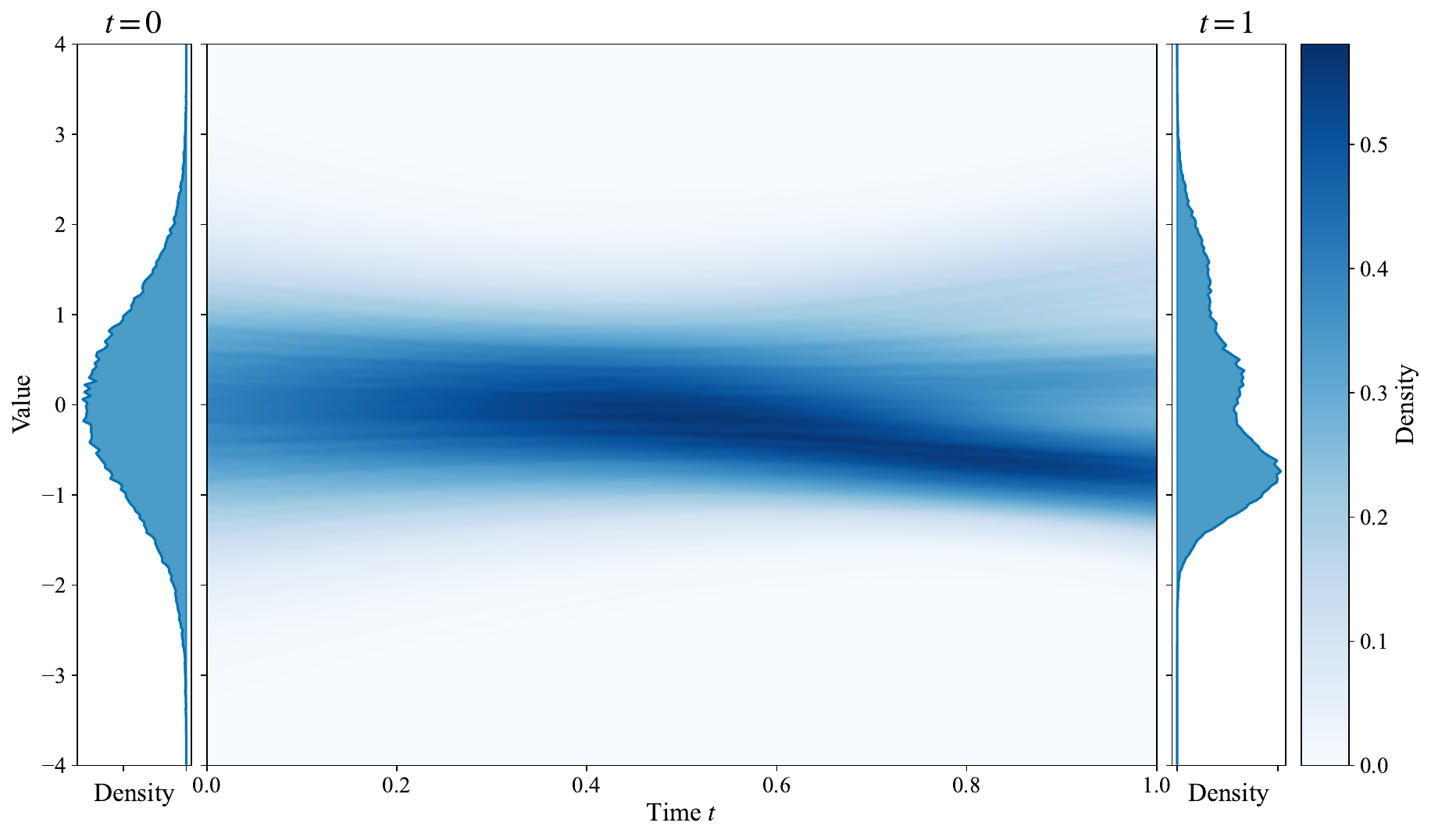}
\caption{An illustration of the CFM generation process. The model learns to transform a simple base Gaussian distribution (left) to a complex three-peak mixed-Gaussian target distribution (right) by learning the intermediate velocity field (middle).\label{fig:flow_trajectory}}
\end{figure}

Flow Matching (FM) formulates the generation process as modeling a time-dependent vector field (also referred to as a velocity field) $v_\theta(x, t)$, whose induced Ordinary Differential Equation (ODE) transports a simple prior probability measure $p_0 = \mathcal{N}(0, I)$ to a complex target data distribution $p_1$. Since directly regressing the marginal vector field is computationally intractable, we adopt Conditional Flow Matching (CFM)~\cite{lipmanFlowMatchingGenerative2022}. By conditioning on individual data samples to construct simple, independent probability paths, CFM provides a tractable regression objective that, in expectation, perfectly recovers the underlying marginal vector field.

Given a target data sample $x_1 \sim p_\mathrm{data}$ and noise $x_0 \sim \mathcal{N}(0, I)$, CFM constructs interpolated samples along a linear conditional path:
\begin{equation}
    x_t = (1 - t) x_0 + t x_1, \quad t \in [0, 1].
\end{equation}
The conditional vector field generating this path is the constant derivative: $u_t = \frac{dx_t}{dt} = x_1 - x_0$. The CFM loss then trains a neural network to regress against this target velocity:
\begin{equation}
    \mathcal{L}_\mathrm{CFM}(\theta) = \mathbb{E}_{t, x_0, x_1} \left[ \| v_\theta(x_t, t) - (x_1 - x_0) \|^2 \right].
\end{equation}
An intuitive visual representation of this learned velocity mapping, transporting samples from a base Gaussian noise state through intermediate trajectories toward a multi-modal deterministic target, is provided in Figure~\ref{fig:flow_trajectory}.

We use two distinct network variants under this shared CFM objective:

\paragraph{Unconditional generation network.} For generative simulation, we train an unconditional velocity network $v_\theta(x_t, t)$ that takes as input the current state $x_t$ concatenated with a Fourier embedding of $t$, autonomously mapping pure Gaussian noise into the target physics distribution.

\paragraph{Conditional unfolding network.} For detector unfolding, we train a conditional velocity network $v_\theta(x_t, t \mid c)$ parameterized by the detector-level measurement $c$. In this context, $c$ explicitly encodes the resolution-degraded kinematic observables---specifically, the Gaussian-smeared spatial momenta of the $\pi^\pm$ tracks representing finite instrumental precision. This approach is conceptually analogous to conditional invertible-network unfolding~\cite{backesUnfoldingMethodBased2024} but utilizes a simulation-free CFM objective. The conditioning vector $c$ is processed through a learned two-layer SiLU-activated Multi-Layer Perceptron (MLP) and concatenated alongside the $x_t$ state and time embedding. During inference, this condition $c$ is held constant throughout the entire ODE integration, allowing the network to deterministically recover particle-level kinematics from localized, smeared observations.

Both variants utilize residual backbone architectures~\cite{heDeepResidualLearning2016} with SiLU activations. At inference, generation is performed by integrating the learned ODE from $t=0$ to $t=1$ via an adaptive Dormand-Prince solver using deterministic paths, establishing a deterministic mapping between the prior and target distributions. Comprehensive hyperparameter configurations of both networks are provided in Appendix~\ref{app:hyperparameters}.

\subsection{Physics-informed metrics}%
\label{subsec:metrics}

To rigorously evaluate model performance, we monitor the following physics-informed metrics in addition to standard training metrics, such as loss, during the validation and testing phases:

\begin{enumerate}
    \item \emph{Marginals:} We report the $\chi^2$ statistic and the Wasserstein-1 distance ($W_1$) between generated and true univariate distributions.
    
    \item \emph{Pairwise joints:} We report the 2D binned $\chi^2$ statistic ($\chi^2_{\mathrm{2D}}$) over all feature pairs, testing whether the model captures bivariate dependencies beyond individual marginals.
    
    \item \emph{Global correlation structure:} We report the correlation matrix distance $D_{\mathrm{corr}} = \|\mathrm{corr}(\text{truth}) - \mathrm{corr}(\text{gen})\|_F$, the Frobenius norm of the Pearson correlation-matrix difference, measuring the holistic reproduction of linear dependencies across all channels.
    
    \item \emph{Memorization:} We report the nearest-neighbor distance ratio $R_{\mathrm{NN}} = \bar{d}_{\mathrm{gen \to train}} / \bar{d}_{\mathrm{train \to train}}$, where $\bar{d}$ denotes the mean $L^2$ nearest-neighbor distance. A ratio $R_{\mathrm{NN}} \approx 1$ indicates generalization, whereas $R_{\mathrm{NN}} \ll 1$ flags memorization.
\end{enumerate}

In tabular summaries, we present $\chi^2$, $W_1$, $\chi^2_{\mathrm{2D}}$, $D_{\mathrm{corr}}$, and $R_{\mathrm{NN}}$. For dynamic tracking (Figure~\ref{fig:loss_vs_physics_metrics}), we monitor training loss, the Number of Function Evaluations (NFE; lower values indicate straighter flows for adaptive solvers), $W_1$, and $D_{\mathrm{corr}}$. Detailed mathematical formulations are in Appendix~\ref{app:metrics}.

\section{Results}%
\label{sec:results}

\subsection{Synthetic benchmark validation}%
\label{subsec:synthetic_results}

To isolate modeling challenges, we first validated the architecture on synthetic 1D distributions; detailed results with complex topologies are provided in Appendix~\ref{app:synthetic}.

\subsection{Performance on MC-POM dataset}%
\label{subsec:mcpom_results}

\begin{figure}[htbp]
\centering
\begin{subfigure}[t]{0.5\textwidth}
\centering
\includegraphics[width=\linewidth]{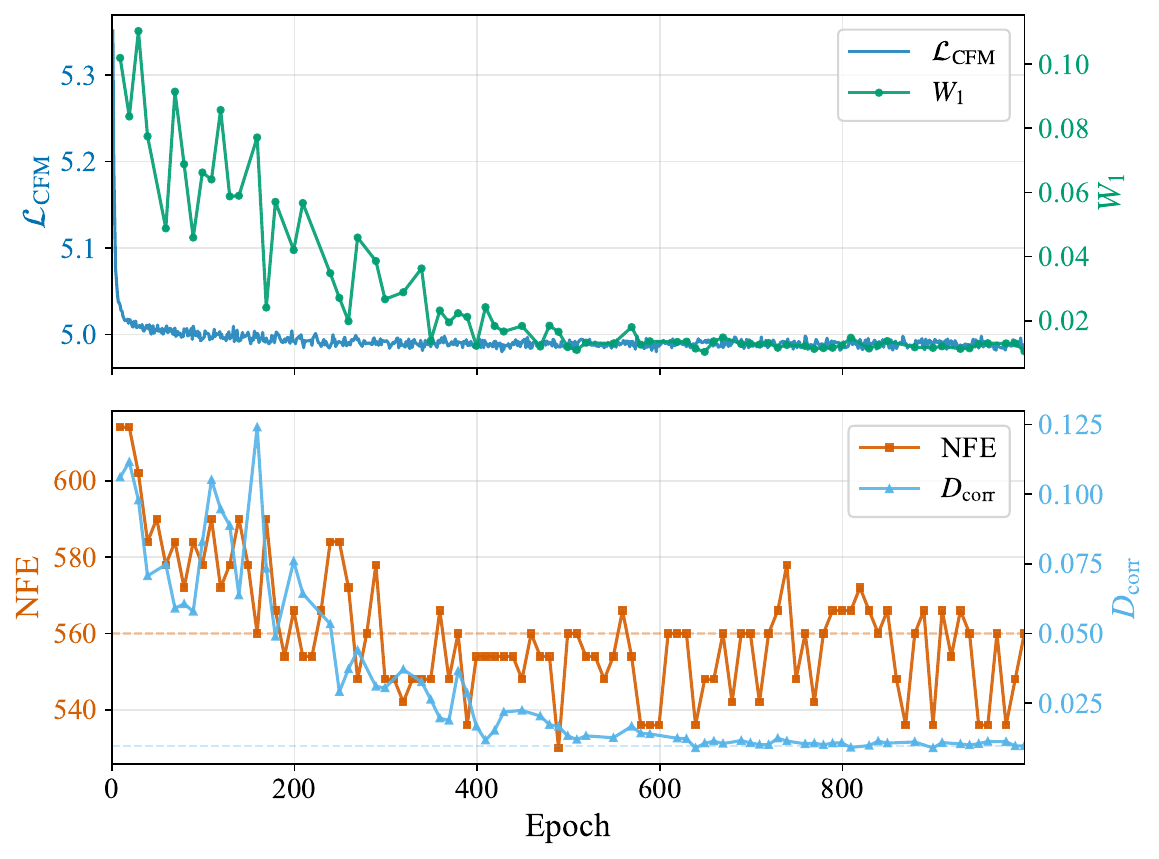}
\caption{\label{fig:loss_vs_physics_metrics}}
\end{subfigure}\hfill
\begin{subfigure}[t]{0.5\textwidth}
\centering
\includegraphics[width=\linewidth]{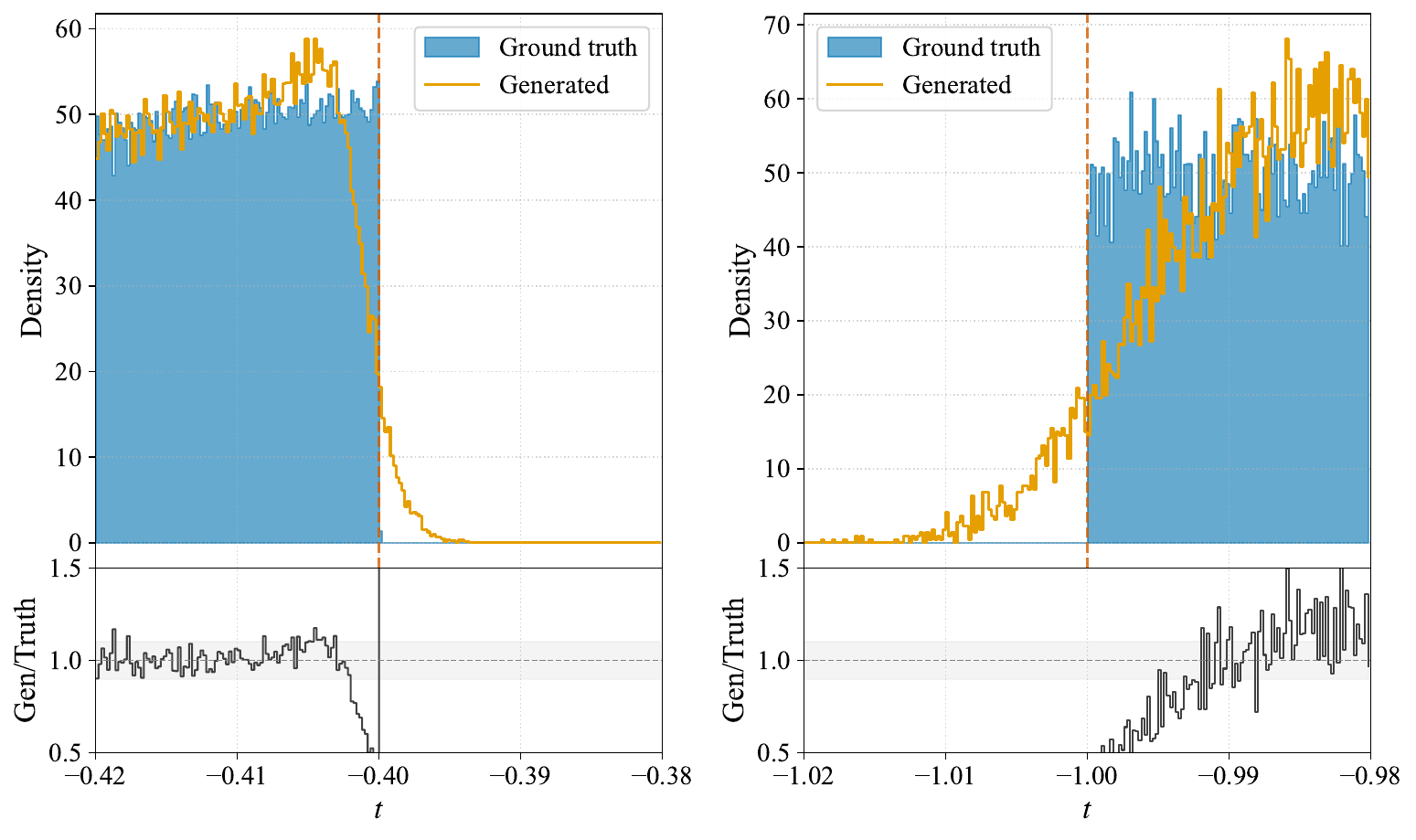}
\caption{\label{fig:mcpom_closeup}}
\end{subfigure}
\caption{Diagnostics on the MC-POM generation task. (a) Training metrics tracked over time, comparing the CFM loss against physics-informed indicators. (b) Close-up comparison of generated and ground-truth distributions in two sharp cut-off regions of the $t$-channel.\label{fig:mcpom_diagnostics}}
\end{figure}

A crucial observation during MC-POM training is that the convergence of the standard CFM velocity loss does not align with true physical fidelity. As Figure~\ref{fig:loss_vs_physics_metrics} shows, the CFM loss plateaus rapidly after ${\sim}20$ epochs. In contrast, physics-informed metrics ($W_1$, $D_{\mathrm{corr}}$), evaluated on the held-out validation split, improve steadily until epoch 600. Thus, CFM loss alone cannot guarantee accurate kinematic reconstruction, necessitating decoupled physical validation metrics.

Table~\ref{tab:mcpom_results} summarizes quantitative performance on the MC-POM dataset for generation and unfolding tasks. All evaluations report metrics from the best-performing checkpoint. Distributional metrics (including the unnormalized 50-bin $\chi^2$, $W_1$, $\chi^2_{\mathrm{2D}}$, $D_{\mathrm{corr}}$) are computed on the held-out test split using 0.8M generated events (matched 1:1 to the test split size), whereas $R_{\mathrm{NN}}$ compares 80K generated events against the 6.4M-event training split to diagnose memorization.

\begin{figure}[htbp]
\centering
\includegraphics[width=\textwidth]{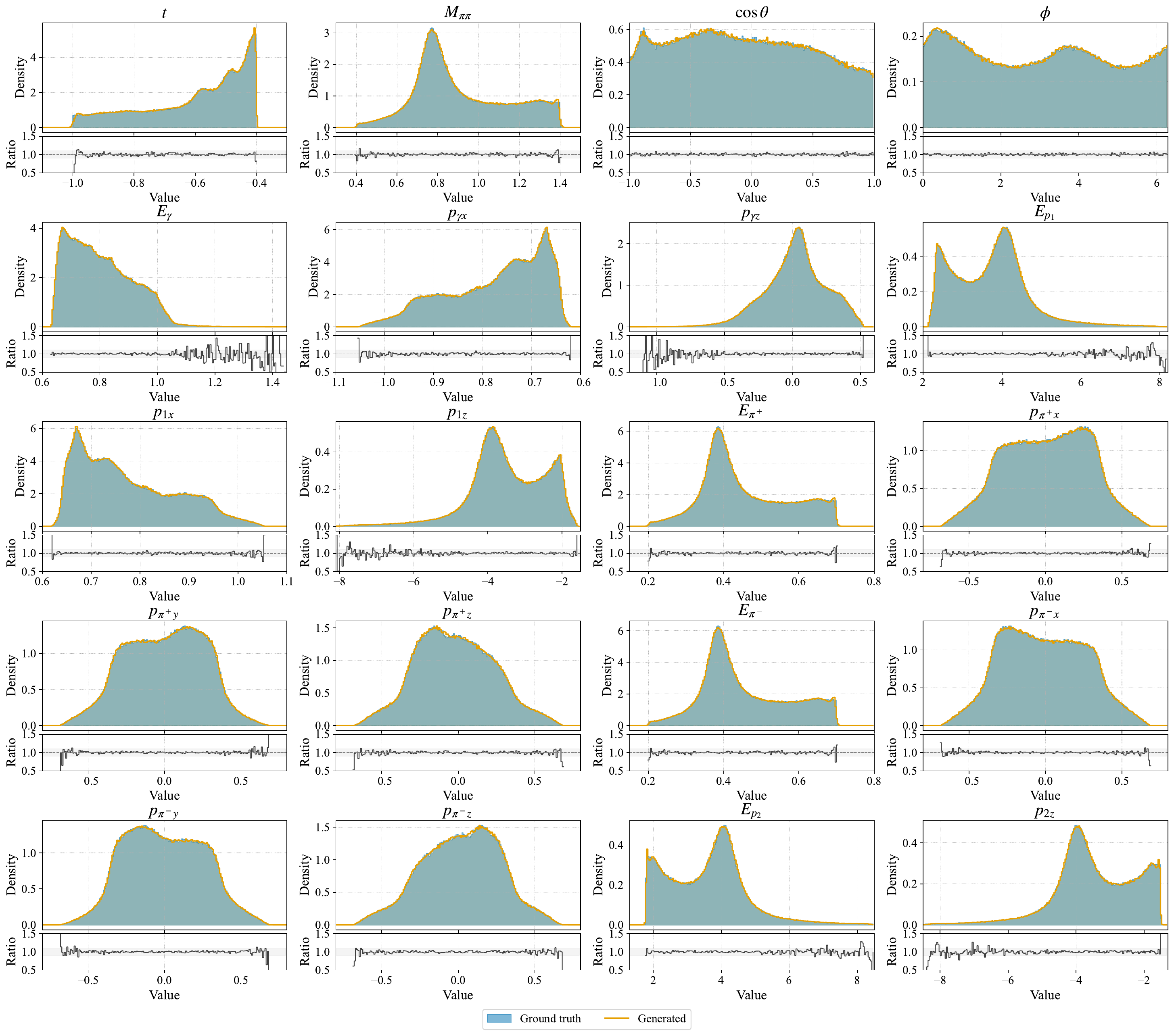}
\caption{Comparison of generated kinematic distributions produced by the CFM model against the ground truth on the JLab MC-POM dataset.\label{fig:mcpom_results}}
\end{figure}

For unconditional generation, the model achieves high-fidelity sampling over the entire phase space, with $R_{\mathrm{NN}} \approx 1.00$ confirming generalization. Figure~\ref{fig:mcpom_results} shows strong agreement between generated and ground-truth distributions (correlation matrix in Appendix~\ref{app:extended_generation}, Figure~\ref{fig:correlation_matrix}). Exact-zero channels and axis units (GeV, GeV$^2$, radians) are omitted for clarity. A close-up of the $t$-channel (Figure~\ref{fig:mcpom_closeup}), computed from the generated 10D momenta, reveals minor deviations only near hard kinematic cutoffs---a known limitation of continuous flows.

\begin{figure}[htbp]
\centering
\includegraphics[width=\textwidth]{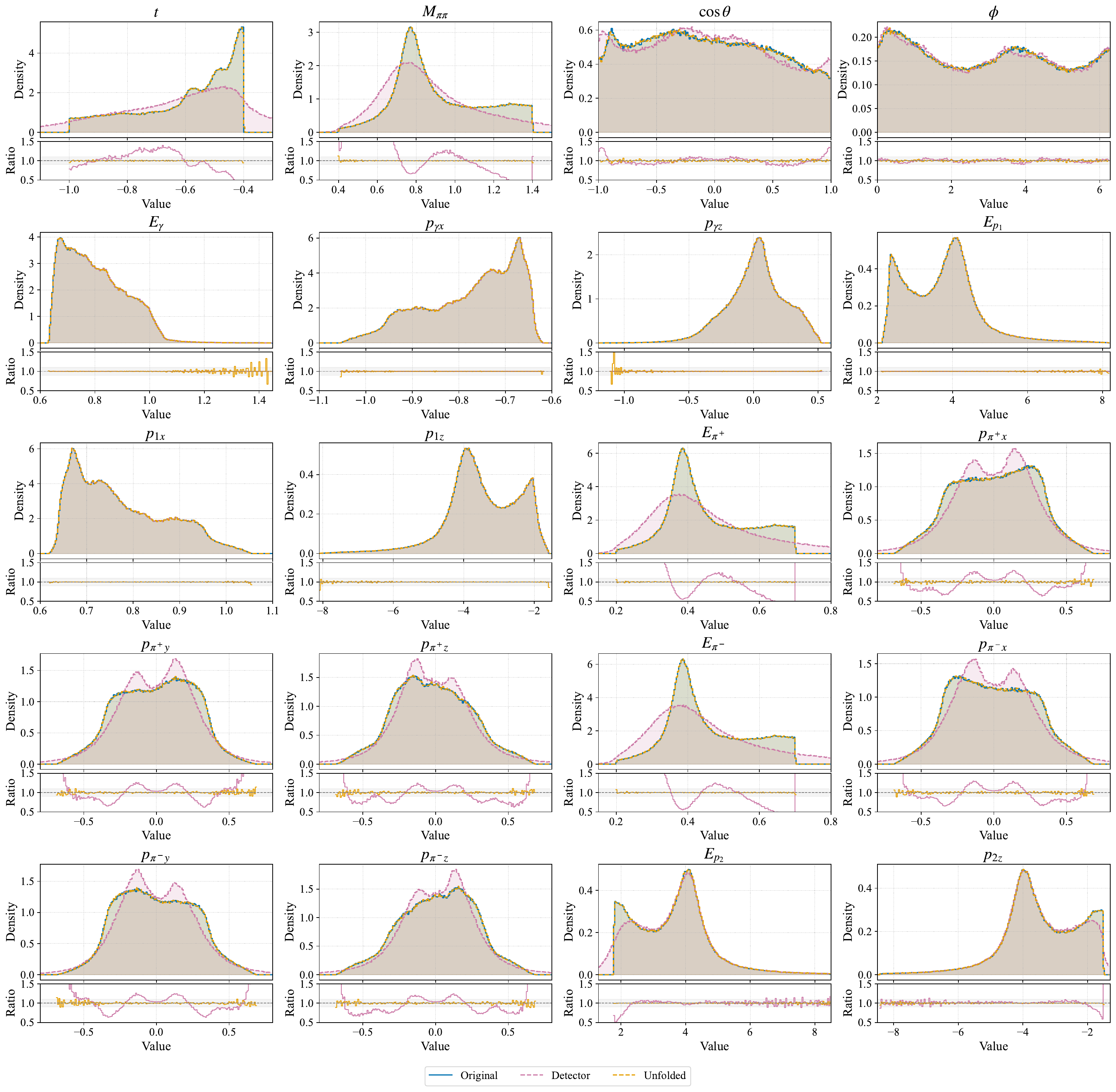}
\caption{Unfolding ($\sigma_{\mathrm{smear}}=1.0$) mapping detector-level distributions back to particle-level truth, compared against the ground truth and smeared detector-level inputs.\label{fig:denoising_results}}
\end{figure}

For detector unfolding, the model deterministically maps smeared observations back to particle-level truth. Table~\ref{tab:mcpom_results} shows unfolding metrics remain comparable across smearing scales; small variations are likely attributable to training stochasticity. Figure~\ref{fig:denoising_results} confirms degraded variables at $\sigma_{\mathrm{smear}}=1.0$ are restored to high-fidelity distributions (extended validations in Appendix~\ref{app:additional_unfolding}). $R_{\mathrm{NN}}$ is omitted for unfolding because proximity to the training manifold is the desired objective there, not a memorization failure mode.

\begin{table}[htbp]
\centering
\caption{MC-POM generation and detector unfolding performance across various smearing scales $\sigma_{\mathrm{smear}}$. A ratio $R_{\mathrm{NN}} \approx 1$ confirms generalization without data memorization.}
\label{tab:mcpom_results}
\begin{tabular}{ll|ccccc}
\toprule
Task & $\sigma_{\mathrm{smear}}$ & $\chi^2\downarrow$ & $W_1\downarrow$ & $\chi^2_\text{2D}\downarrow$ & $D_{\mathrm{corr}}\downarrow$ & $R_{\mathrm{NN}}$ \\
\midrule
Generation & --- & 33503.2 & $8.96 \times 10^{-4}$ & 33962.4 & 1.53 & 1.01 \\
\midrule
\multirow{3}{*}{Unfolding} & 2.0 & 33364.4 & $1.51 \times 10^{-4}$ & 29245.0 & 1.44 & --- \\
 & 1.0 & 33355.6 & $1.60 \times 10^{-4}$ & 40226.7 & 1.48 & --- \\
 & 0.5 & 33346.0 & $1.28 \times 10^{-4}$ & 38270.9 & 2.11 & --- \\
\bottomrule
\end{tabular}
\end{table}

\section{Discussion and conclusion}%
\label{sec:discussion}

For practitioners applying generative models across NP and HEP, our results show that standard CFM velocity loss convergence can mislead and does not ensure physical-observable fidelity. Because physical distributions need longer training to stabilize, decoupled evaluation using domain-specific observables is essential. ScatterPrism---validated on synthetic and JLab datasets---demonstrates that CFM provides a robust, deterministic framework for capturing phase spaces and unfolding detector kinematics.

By learning flexible mappings between detector observables and particle truths, ScatterPrism provides an AI-driven, unbinned alternative to expensive Monte Carlo simulations. Initiated by the convergence pathologies exposed in the JLab NP dataset, our methodology reliably captures low cross-section topologies (e.g., near-threshold $J/\psi$ photoproduction, exotic mesons) vital for the EIC. Furthermore, our synthetic benchmarks demonstrate that this diagnostic framework is inherently dataset-agnostic. Consequently, identical machinery is positioned to extend to complex HEP final states, such as jets and high-multiplicity multi-particle decays. This unified approach enables rapid systematic iterations without repeated simulation campaigns (see computational throughput in Appendix~\ref{app:inference_speed}).

ScatterPrism prioritizes modularity and reproducibility via PyTorch Lightning~\cite{falconPyTorchLightning2026}, Hydra~\cite{FacebookresearchHydra2026}, and Weights \& Biases~\cite{wandb}. We utilized standard CFM, achieving high-fidelity structural reconstruction across joints and marginals without the high cost of calculating optimal couplings via Optimal Transport Conditional Flow Matching (OT-CFM) on large datasets.

Beyond nuclear physics, this methodology naturally extends to domains such as molecular dynamics, medical imaging, and astrophysics~\cite{islamCosmo3DFlowWaveletFlow2026, xiaPyTorchFireGPUacceleratedWildfire2025}, which frequently require mapping noisy observations back to foundational truth—mirroring detector unfolding. Our multi-metric validation provides a blueprint for ensuring generative models satisfy physical constraints rather than merely minimizing probability divergences.

Having established the necessity of physics-informed convergence diagnostics in CFM, future deployments will natively integrate GEANT-based detector simulations. The lightweight architecture supports uncertainty quantification, architectural ablations, and benchmarking. Future pipelines will add tests for out-of-distribution generalization and explicit physics-informed loss functions~\cite{bastekPhysicsInformedDiffusionModels2025, baldanPhysicsVsDistributions2025, tauberschmidtPhysicsConstrainedFineTuningFlowMatching2025} to eliminate invalid generations and improve unfolding precision.

\acknowledgments%
We thank Xuweiyi Chen (University of Virginia) for sharing his valuable experience in developing the model. We also thank Huilin Huang (University of Virginia) for her financial support. This work was partially supported by the National Science Foundation under POSE award 2346173.

\section*{Data and Code Availability}%
Code, models, and datasets are available on GitHub (\url{https://github.com/xiazeyu/ScatterPrism}) and Zenodo (\url{https://doi.org/10.5281/zenodo.20407373}, \url{https://doi.org/10.5281/zenodo.20391126}).

\section*{Artificial Intelligence Disclosure}%
The authors utilized Gemini~3.1~Pro, Claude~Opus~4.5/4.6/4.7 to refine prose and assist with code/documentation. All AI-generated content was thoroughly reviewed, verified, and edited. The authors take full responsibility for the content, accuracy, and integrity of this publication.

\appendix

\section{Network architecture and training details}
\label{app:hyperparameters}

Building upon the methodology in Section~\ref{subsec:cfm_framework}, both network variants share a residual MLP architecture. Table~\ref{tab:network_arch} lists the primary hyperparameters. The unconditional network evaluates the concatenated state $[x_t \,\|\, e_t] \in \mathbb{R}^{D + d_t}$, where $e_t$ is the Fourier time embedding. For the conditional variant, the detector measurement $c \in \mathbb{R}^{D}$ is embedded into $e_c \in \mathbb{R}^{128}$, expanding the input to $[x_t \,\|\, e_t \,\|\, e_c] \in \mathbb{R}^{D + d_t + 128}$.

\begin{table}[htbp]
\centering
\caption{Network architecture and training hyperparameters.}
\label{tab:network_arch}
\begin{tabular}{ll|ll}
\toprule
\textbf{Architecture} & \textbf{Value} & \textbf{Training} & \textbf{Value} \\
\midrule
Input dimension $D$ & 10 & Optimizer & AdamW \\
Hidden dimensions & $[512]^6$ (6 layers) & Learning rate & $10^{-4}$ (decay 0.5, patience 50) \\
Time embedding dim. & 64 & Weight decay & $10^{-5}$ \\
Activation function & SiLU & Batch size & 20K \\
Total parameters & $\sim$2.7M & Total epochs & 1000 \\
\bottomrule
\end{tabular}
\end{table}

\paragraph{Time conditioning.} The scalar time $t \in [0, 1]$ is embedded using 32 geometrically-spaced frequencies up to a maximum frequency $\omega_{\max}=64$. A learned linear projection maps this Fourier feature into a $\mathbb{R}^{64}$ vector, providing the network with high-frequency temporal components necessary for resolving sharp changes in the velocity field.

\paragraph{Residual blocks.} The six hidden widths in Table~\ref{tab:network_arch} expand into an input linear projection, followed by five residual blocks, followed by an output linear projection. Each residual block applies a two-layer SiLU-activated sequence, $\text{SiLU}\bigl(\text{Linear}(\text{SiLU}(\text{Linear}(h)))\bigr) + h$, promoting stable gradient flow and facilitating the learning of identity mappings where the vector field is approximately constant.

\paragraph{Optimization details.} We optimize parameters with \texttt{AdamW}. A \texttt{ReduceLROnPlateau} scheduler tracks the epoch-level validation CFM loss (\texttt{val/loss\_epoch}), decaying the initial $10^{-4}$ learning rate by a factor of 0.5 upon detecting plateaus over a 50-epoch patience threshold (capped minimally at $10^{-7}$). For efficiency, validation metrics are computed each epoch on the full held-out validation split using a fixed-step fourth-order Runge--Kutta (RK4) integrator to accelerate the training loop; the adaptive Dormand--Prince (DOPRI5) solver described below is used for all final-prediction reporting. The NFE traces in Figure~\ref{fig:loss_vs_physics_metrics} are recomputed post hoc with DOPRI5 on a fixed 50K-event subset per checkpoint. The globally optimal checkpoint persisting into final tabular summaries monitors the pure generative physical reconstruction (\texttt{val/chi2\_mean}) over the raw CFM velocity loss, as discussed in detail throughout Section~\ref{subsec:mcpom_results}.

\paragraph{Generation and inference details.} For the Dormand-Prince (DOPRI5) ODE solver introduced in Section~\ref{subsec:cfm_framework}, absolute and relative tolerances are uniformly set to $10^{-7}$ for unconditional generation and relaxed to $10^{-3}$ for unfolding tasks. On a single NVIDIA RTX A6000 GPU, one training iteration on a 20K-event batch completes in $\sim$24\,ms ($\sim$820K samples/s). Inference throughput reaches $\sim$3.0K samples/s for strict unconditional generation and $\sim$83K samples/s for conditional unfolding, with a detailed breakdown provided in Section~\ref{app:inference_speed}.

\paragraph{Implementation and environment details.}
The ScatterPrism framework is implemented in Python ($\ge$3.13) utilizing PyTorch ($\ge$2.10.0) and PyTorch Lightning (2.6.4)~\cite{falconPyTorchLightning2026} for hardware-agnostic training. Core underlying dependencies include \texttt{torchdyn} (1.0.6) for numerical ODE integration. The V100 node used PyTorch 2.10.0 with CUDA 12.8, while the A6000 and dedicated CPU nodes used PyTorch 2.12.0 with CUDA 13.0.

All model training and computational evaluations were primarily executed on the UVA Rivanna High-Performance Computing Cluster. GPU acceleration was performed using a single NVIDIA Tesla V100-SXM2 tensor core GPU (32\,GB VRAM; utilizing 4 cores of an Intel Xeon Gold 6230 @ 2.10\,GHz, SMT off) and an NVIDIA RTX A6000 GPU (48\,GB VRAM; utilizing 4 cores of an AMD EPYC 7352, SMT off). CPU-only benchmarking was performed on a dedicated dual-socket AMD EPYC 9454 node (2$\times$48 physical cores with SMT disabled, 96 cores total). Exact SLURM submission scripts and environment configurations used for these benchmarks are provided in the accompanying artifacts.

\section{Evaluation metrics}
\label{app:metrics}

To rigorously assess the fidelity of the generated kinematic distributions against the ground truth, we utilize a structured set of evaluation metrics spanning marginal distributions, multivariate structures, and network memorization characteristics:

\paragraph{Marginals.}
The quality of individual feature distributions is measured primarily using the $\chi^2$ statistic and the Wasserstein-1 distance ($W_1$):
\begin{itemize}
    \item \textbf{Mean $\chi^2$ (\texttt{val/chi2\_mean}):} Evaluated over 1D binned histograms of individual features. To prevent artifacts from outlier limits, exactly 50 uniform bins are dynamically bound between the minimum and maximum values of the true expected distribution. The generated histograms are normalized to match the total event count of the truth distribution before evaluation. Formally, for a single feature, the statistic is:
    \begin{equation}
        \chi^2 = \sum_{i=1}^{50} \frac{(O_i - E_i)^2}{E_i}
    \end{equation}
    where $O_i$ is the normalized generated count and $E_i$ is the expected true count in the $i$-th bin. 
    \item \textbf{Mean Wasserstein distance (\texttt{val/wasserstein\_mean}):} Measures the minimum mass-transport distance required to transform the generated 1D marginal distributions to the true distributions. To explicitly circumvent the manual binning limits of $\chi^2$, this metric evaluates the raw, unbinned 1D distributions via their cumulative distribution functions (CDFs) $F(x)$ and $G(x)$:
    \begin{equation}
        W_1 = \int_{-\infty}^{\infty} |F(x) - G(x)| \, dx
    \end{equation}
\end{itemize}

\paragraph{Pairwise joints.}
To evaluate bivariate dependencies between kinematic variables, we examine two-dimensional distributions:
\begin{itemize}
    \item \textbf{Mean 2D $\chi^2$ (\texttt{val/chi2\_2d\_mean}):} Extends the $\chi^2$ formulation to two-dimensional cross-sections across all 45 unique pairwise feature combinations, returning the arithmetic mean. Each axis is divided into 20 uniform bins (yielding exactly 400 rectangular bins per pair), determined entirely by the ground-truth coordinate extremes. This rigorously tests whether the generator captures underlying multivariate dependencies beyond independent marginals.
\end{itemize}

\paragraph{Global correlation structure.}
To verify that generated events accurately reproduce multi-dimensional kinematic constraints, we assess holistic structural fidelity using:
\begin{itemize}
    \item \textbf{Correlation matrix distance $D_{\mathrm{corr}}$ (\texttt{val/correlation\_distance}):} A holistic measure of how effectively the network reconstructs global linear relationships. It is evaluated as the Frobenius norm of the difference between the sample Pearson correlation matrices: 
    \begin{equation}
        D_{\mathrm{corr}} = \|\mathrm{corr}(\text{truth}) - \mathrm{corr}(\text{gen})\|_F
    \end{equation}
\end{itemize}

\paragraph{Memorization.}
We test for generative generalization using comparative $L^2$ nearest-neighbor mappings:
\begin{itemize}
    \item \textbf{Nearest-neighbor distance ratio $R_{\mathrm{NN}}$ (\texttt{nn/memorization\_ratio}):} Serves as our principal guard against model over-fitting. It calculates the ratio of the mean nearest-neighbor distance from generated samples to the training set ($\bar{d}_{\mathrm{gen \to train}}$) versus the mean nearest-neighbor distance natively found within the training set itself ($\bar{d}_{\mathrm{train \to train}}$):
    \begin{equation}
        R_{\mathrm{NN}} = \frac{\bar{d}_{\mathrm{gen \to train}}}{\bar{d}_{\mathrm{train \to train}}}
    \end{equation}
    A ratio $R_{\mathrm{NN}} \approx 1$ indicates high-fidelity generalization, whereas $R_{\mathrm{NN}} \ll 1$ flags strict dataset memorization. Notably, for certain synthetic mock datasets, the native distance $\bar{d}_{\mathrm{train \to train}}$ can be exceptionally small, causing $R_{\mathrm{NN}}$ to appear abnormally large; this is an expected geometric artifact rather than an indication of model collapse.
\end{itemize}

Additional auxiliary scalars monitored behind the $R_{\mathrm{NN}}$ evaluation include:
\begin{itemize}
    \item \textbf{\texttt{nn/D\_gen\_to\_train\_mean}:} The arithmetic mean $L^2$ Euclidean distance from generated events to their single closest element in the true training manifold.
    \item \textbf{\texttt{nn/D\_train\_to\_train\_mean}:} The native baseline mean $L^2$ distance computed by matching points from a random sub-sample of the training set to the remainder of the training elements.
    \item \textbf{\texttt{nn/D\_gen\_to\_train\_min}} and \textbf{\texttt{nn/D\_train\_to\_train\_min}:} The corresponding absolute minimum scalar values computed for the distributions above.
\end{itemize}

\section{Synthetic benchmark analysis}
\label{app:synthetic}

To isolate and understand the fundamental generative capabilities of the architecture independent of physical kinematics, we conducted extensive evaluations on a suite of synthetic 1D mock datasets. These benchmarks are specifically designed to stress-test the model against complex topological structures commonly encountered in physics, such as multi-modal overlapping resonances, sharp kinematic cut-offs, and high-frequency perturbative noise.

Each synthetic dataset uses the same 8M-event corpus and 8:1:1 train/val/test split as MC-POM (Section~\ref{subsec:datasets}). $\chi^2$ and $W_1$ are computed on the 0.8M-event test split, and $R_{\mathrm{NN}}$ compares 80K generated events against the 6.4M-event training split. Table~\ref{tab:mock_results} summarizes the resulting fidelity and memorization metrics across these diverse 1D topologies. Across the majority of datasets, the model achieves exceptionally low Wasserstein ($W_1$) distances and minimal $\chi^2$ deviations, indicating robust macro-level distribution reconstruction. Crucially, the nearest-neighbor diagnostics confirm that the framework generalizes without duplicating training events: $R_{\mathrm{NN}}$ stays within a narrow band of $0.54$ (\textsc{Narrow-Wide-Overlap}) to $15.39$ (\textsc{Exponential-Decay}) across all realistic topologies, with $D_{g \to t}$ remaining the same order of magnitude as the baseline $D_{t \to t}$. Values mildly above unity indicate generated samples sit slightly farther from any training point than the typical training neighbor---consistent with smooth generalization rather than memorization (where $R_{\mathrm{NN}} \ll 1$). The single large positive outlier, $R_{\mathrm{NN}} \approx 3650$ for \textsc{Uniform-Flat}, is a denominator-driven artifact: a perfectly flat training density packs neighboring events so closely that $D_{t \to t}$ approaches zero, inflating the ratio independent of any model failure.

\begin{table}[htbp]
\centering
\caption{Quantitative evaluation on synthetic 1D benchmarks. The table reports marginal fidelity ($\chi^2$, $W_1$) and memorization characteristics ($D_{g \to t}$, $D_{t \to t}$, $R_{\mathrm{NN}}$) across diverse topological structures.}
\label{tab:mock_results}
\resizebox{\textwidth}{!}{
\begin{tabular}{lrccccc}
\toprule
Dataset & Epoch & $\chi^2\downarrow$ & $W_1\downarrow$ & $D_{g \to t}$ & $D_{t \to t}$ & $R_{\mathrm{NN}}$ \\
\midrule
\textsc{Bimodal-Asym} & 999 & 117.8 & $1.66 \times 10^{-3}$ & $6.65 \times 10^{-7}$ & $2.01 \times 10^{-7}$ & 3.31 \\
\textsc{Exponential-Decay} & 339 & 88.7 & $3.23 \times 10^{-3}$ & $1.58 \times 10^{-5}$ & $1.03 \times 10^{-6}$ & 15.39 \\
\textsc{Gauss-Cutoff} & 639 & 73.8 & $7.38 \times 10^{-4}$ & $8.73 \times 10^{-6}$ & $1.20 \times 10^{-6}$ & 7.28 \\
\textsc{Narrow-Wide-Overlap} & 449 & 89.6 & $6.35 \times 10^{-3}$ & $5.30 \times 10^{-7}$ & $9.74 \times 10^{-7}$ & 0.54 \\
\textsc{Noise-3Spikes} & 789 & 97.7 & $2.20 \times 10^{-3}$ & $5.74 \times 10^{-7}$ & $3.56 \times 10^{-7}$ & 1.61 \\
\textsc{Noise-10Spikes} & 439 & 91.0 & $3.89 \times 10^{-3}$ & $8.05 \times 10^{-7}$ & $5.64 \times 10^{-7}$ & 1.43 \\
\textsc{Tall-Flat-Far} & 539 & 95.0 & $5.27 \times 10^{-3}$ & $2.76 \times 10^{-7}$ & $2.37 \times 10^{-7}$ & 1.17 \\
\textsc{Triple-Flat-Spread} & 519 & 160.6 & $1.15 \times 10^{-2}$ & $2.74 \times 10^{-7}$ & $1.34 \times 10^{-7}$ & 2.04 \\
\textsc{Triple-Mixed} & 389 & 118.5 & $8.36 \times 10^{-3}$ & $1.59 \times 10^{-6}$ & $1.91 \times 10^{-7}$ & 8.35 \\
\textsc{Uniform-Flat} & 399 & 101.2 & $3.05 \times 10^{-3}$ & $1.23 \times 10^{-5}$ & $3.36 \times 10^{-9}$ & 3650.36$^{\ast}$ \\
\bottomrule
\end{tabular}
}
\begin{flushleft}
{\footnotesize $^{\ast}$Denominator-driven artifact: the perfectly uniform training density packs neighboring events so densely that $D_{t \to t} \to 0$, inflating $R_{\mathrm{NN}}$. The model generalizes correctly---this does not indicate memorization.}
\end{flushleft}
\end{table}

\begin{figure}[htbp]
\centering
\begin{minipage}{0.49\textwidth}
\centering
\includegraphics[width=\linewidth]{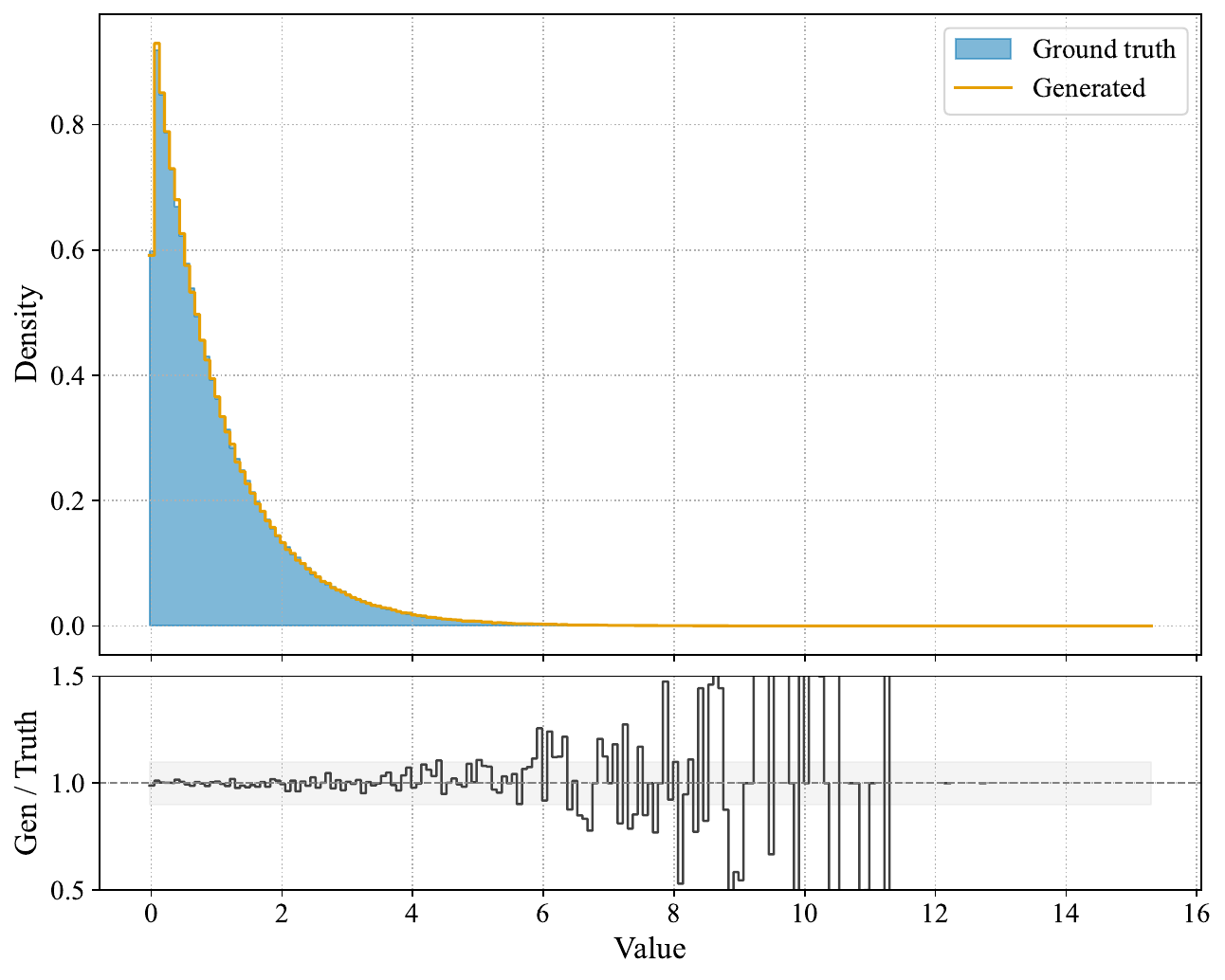}
\caption{Accurate representation of a sharp exponential decay distribution, smoothly capturing the steep initial drop-off.\label{fig:exponential_decay_result}}
\end{minipage}\hfill
\begin{minipage}{0.49\textwidth}
\centering
\includegraphics[width=\linewidth]{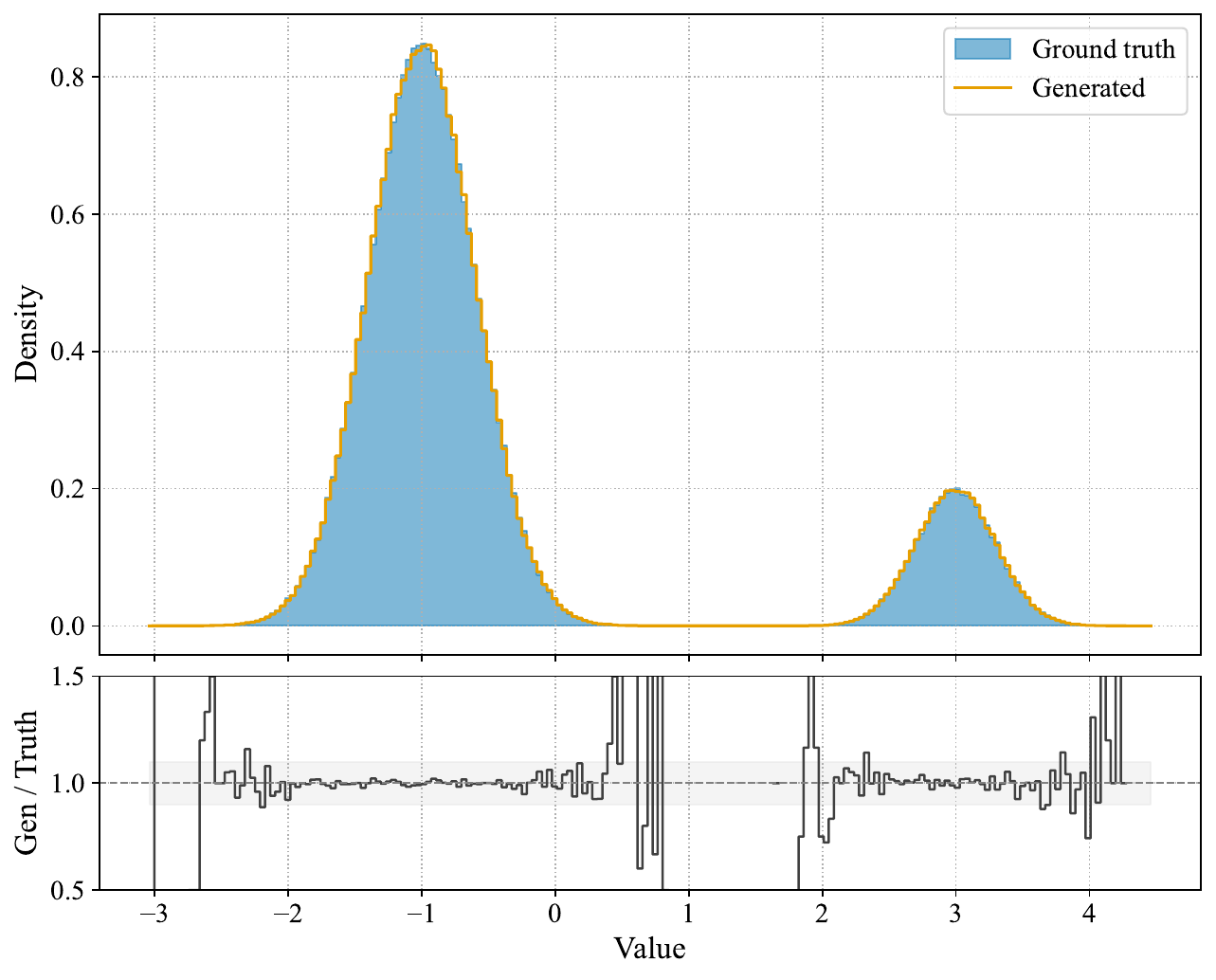}
\caption{High-fidelity reconstruction of an asymmetric bimodal topology, preserving the distinct independent peak heights.\label{fig:bimodal_asym_result}}
\end{minipage}

\vspace{1em}

\begin{minipage}{0.49\textwidth}
\centering
\includegraphics[width=\linewidth]{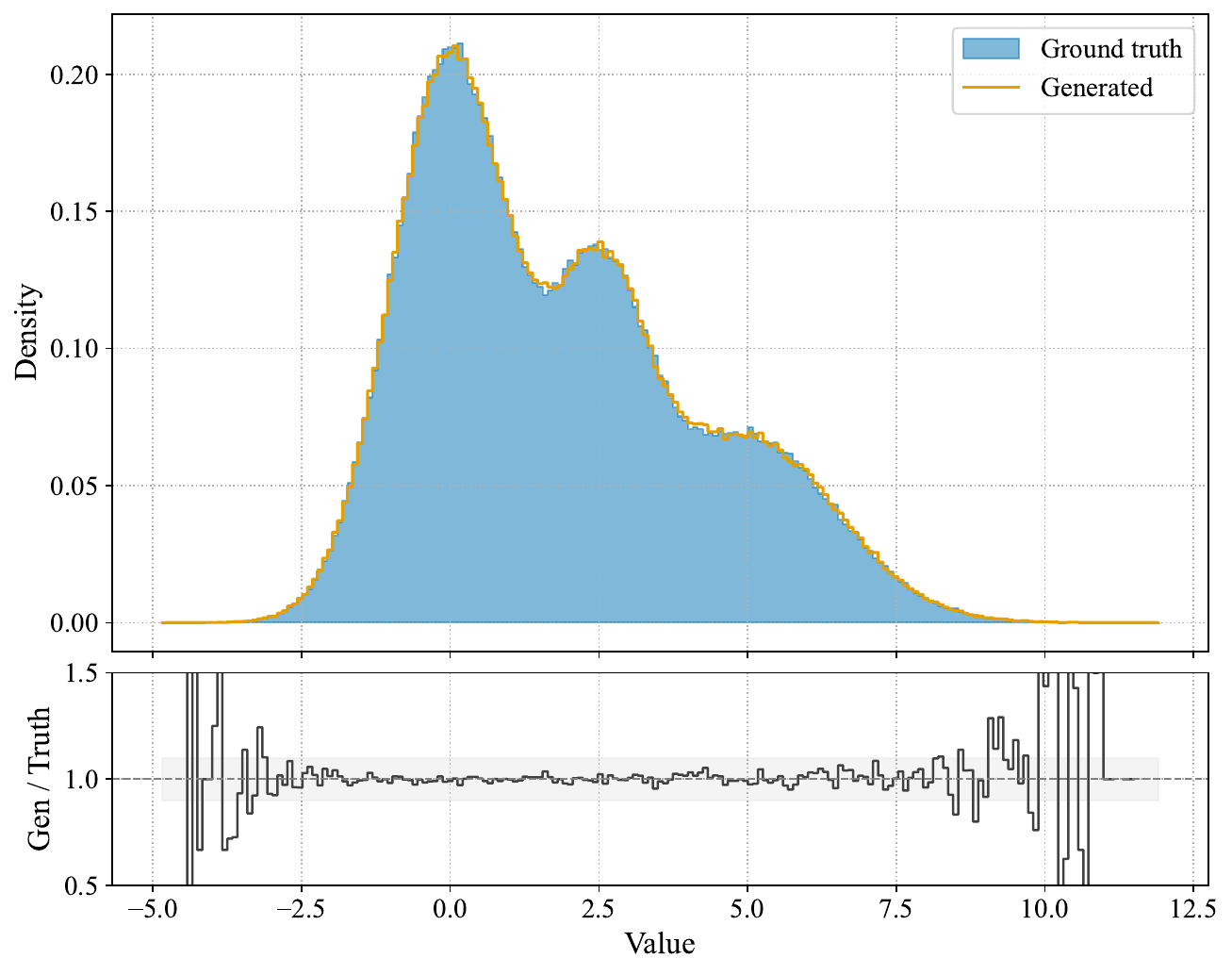}
\caption{Morphological reproduction of a three-peak mixed Gaussian distribution, matching strict boundaries.\label{fig:triple_mixed_result}}
\end{minipage}\hfill
\begin{minipage}{0.49\textwidth}
\centering
\includegraphics[width=\linewidth]{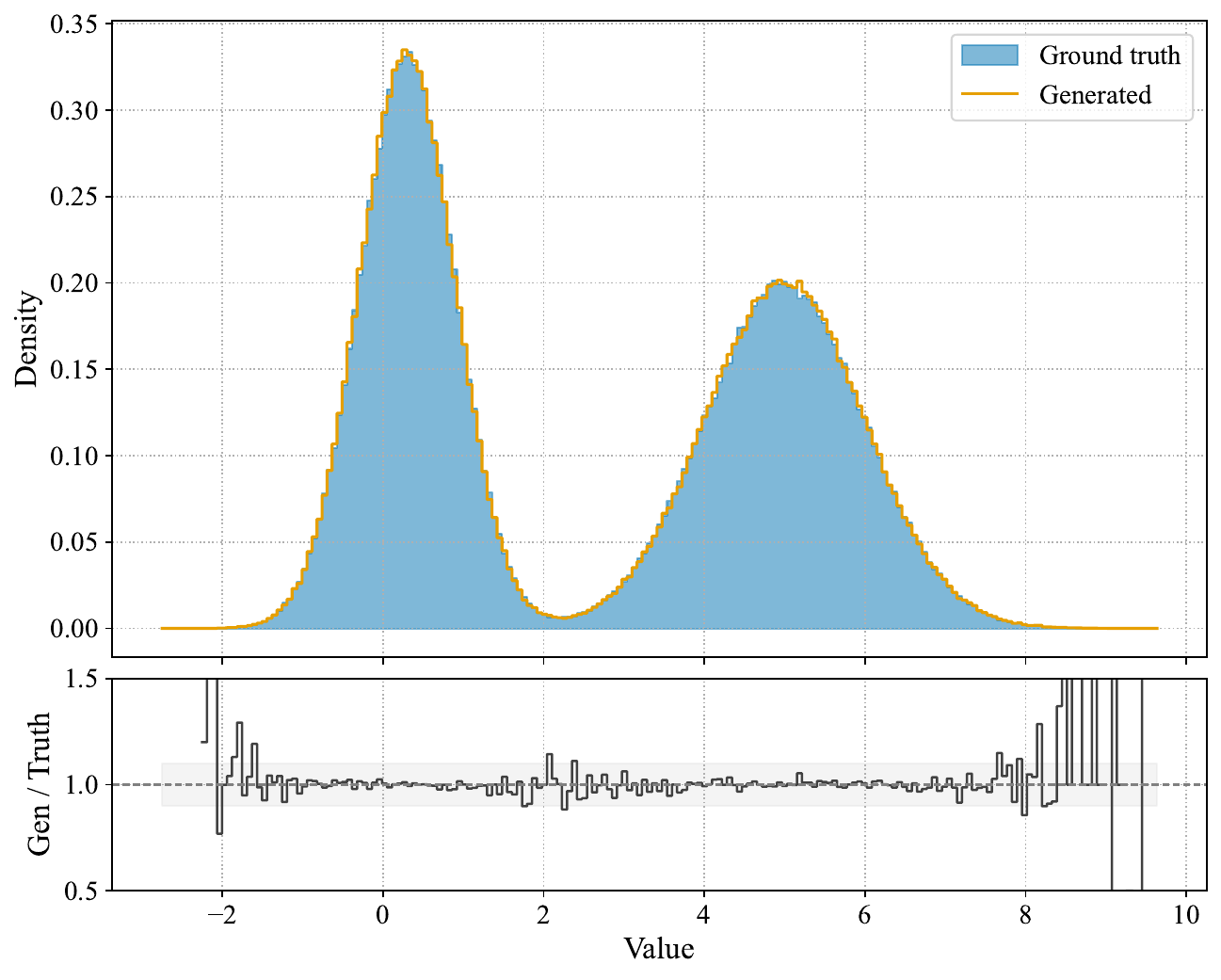}
\caption{Reconstruction of widely separated multi-modal peaks with disparate amplitudes in the tall-flat-far dataset.\label{fig:tall_flat_far_result}}
\end{minipage}
\end{figure}

\begin{figure}[htbp]
\centering
\begin{minipage}{0.49\textwidth}
\centering
\includegraphics[width=\linewidth]{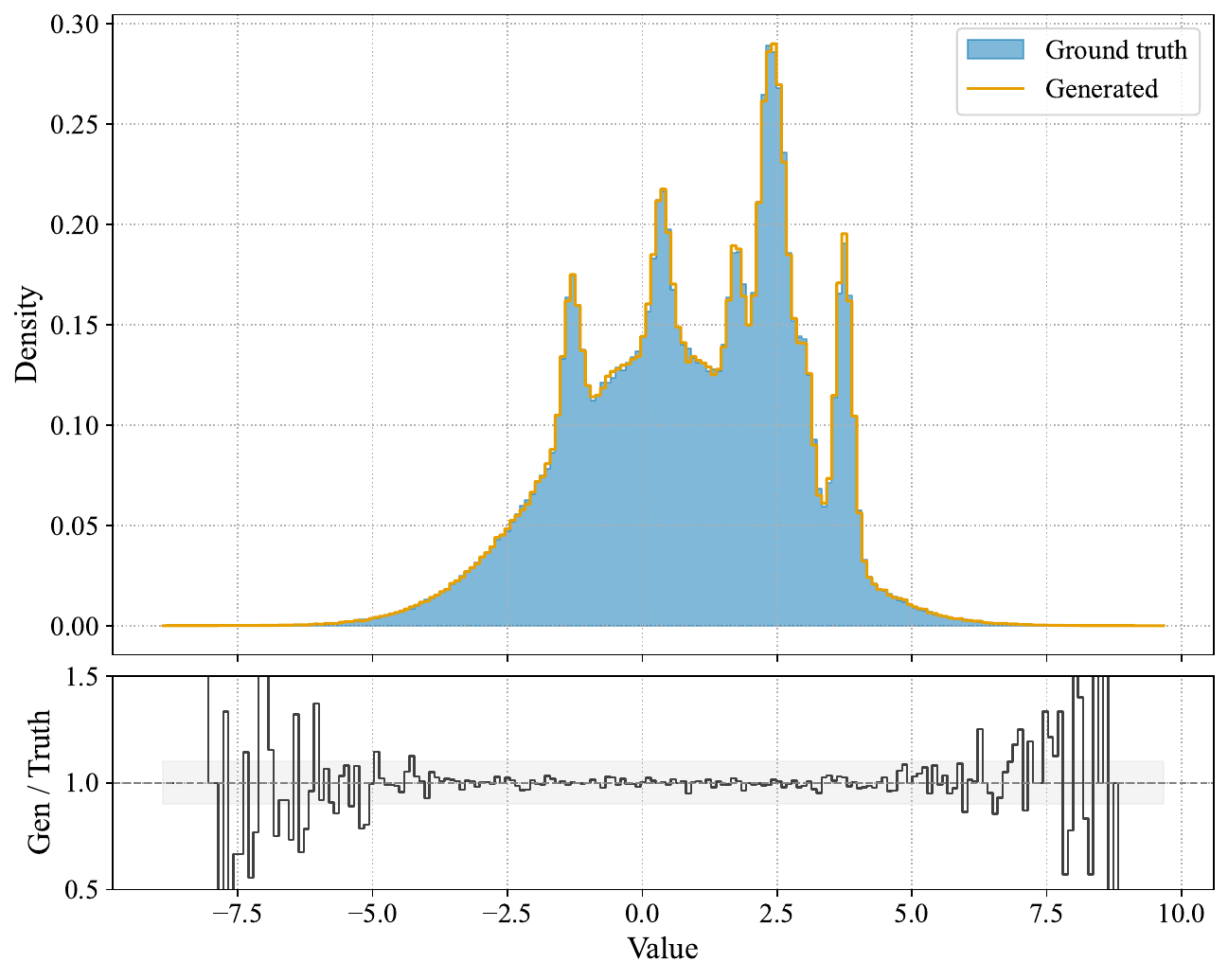}
\caption{Detailed capture of high-frequency noise spikes, successfully resolving dense structural perturbations.\label{fig:noise_10spikes_result}}
\end{minipage}\hfill
\begin{minipage}{0.49\textwidth}
\centering
\includegraphics[width=\linewidth]{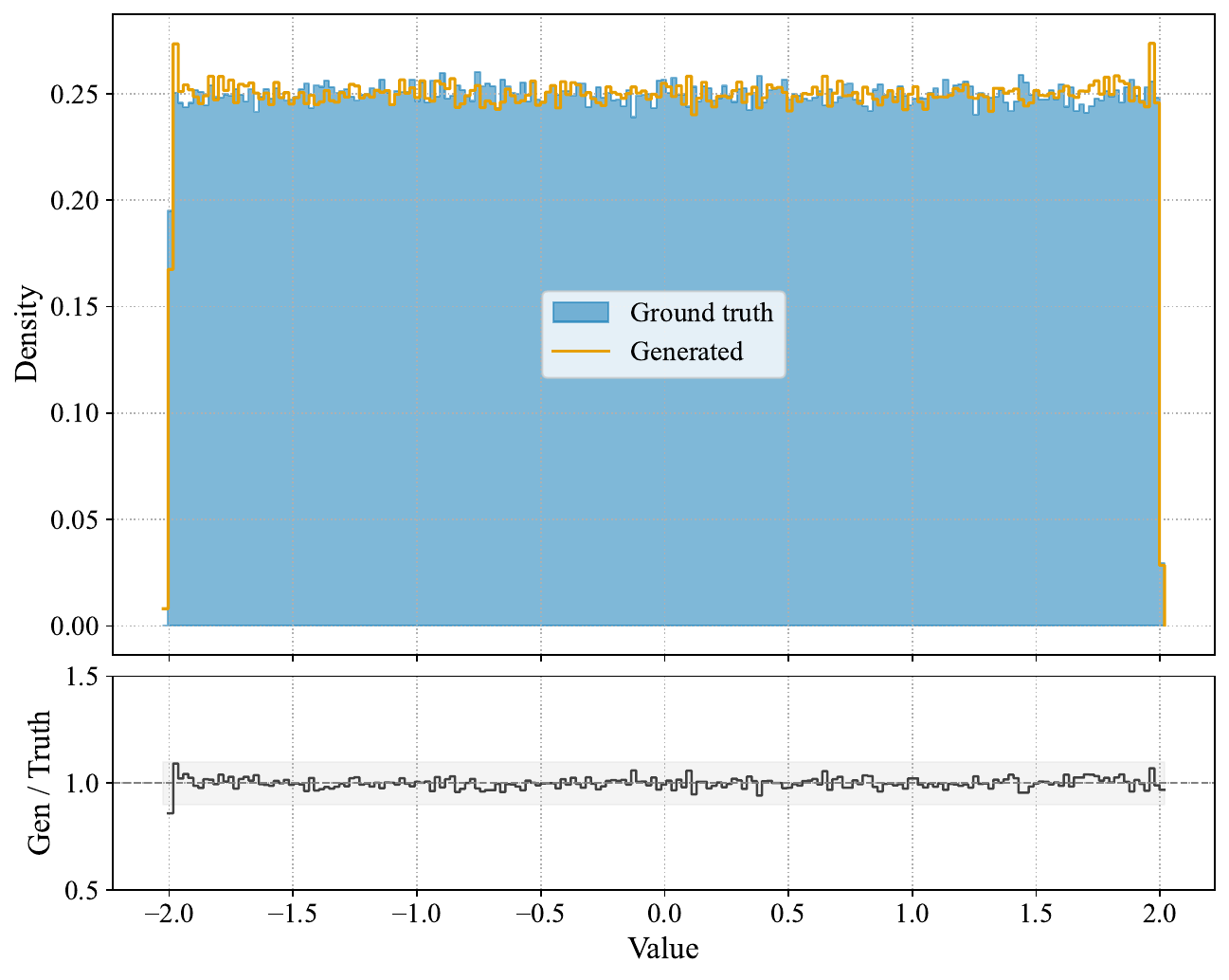}
\caption{Generation on the uniform-flat distribution, illustrating boundary smearing at sharp discontinuous density edges.\label{fig:uniform_flat_result}}
\end{minipage}
\end{figure}

\begin{figure}[htbp]
\centering
\includegraphics[width=0.8\textwidth]{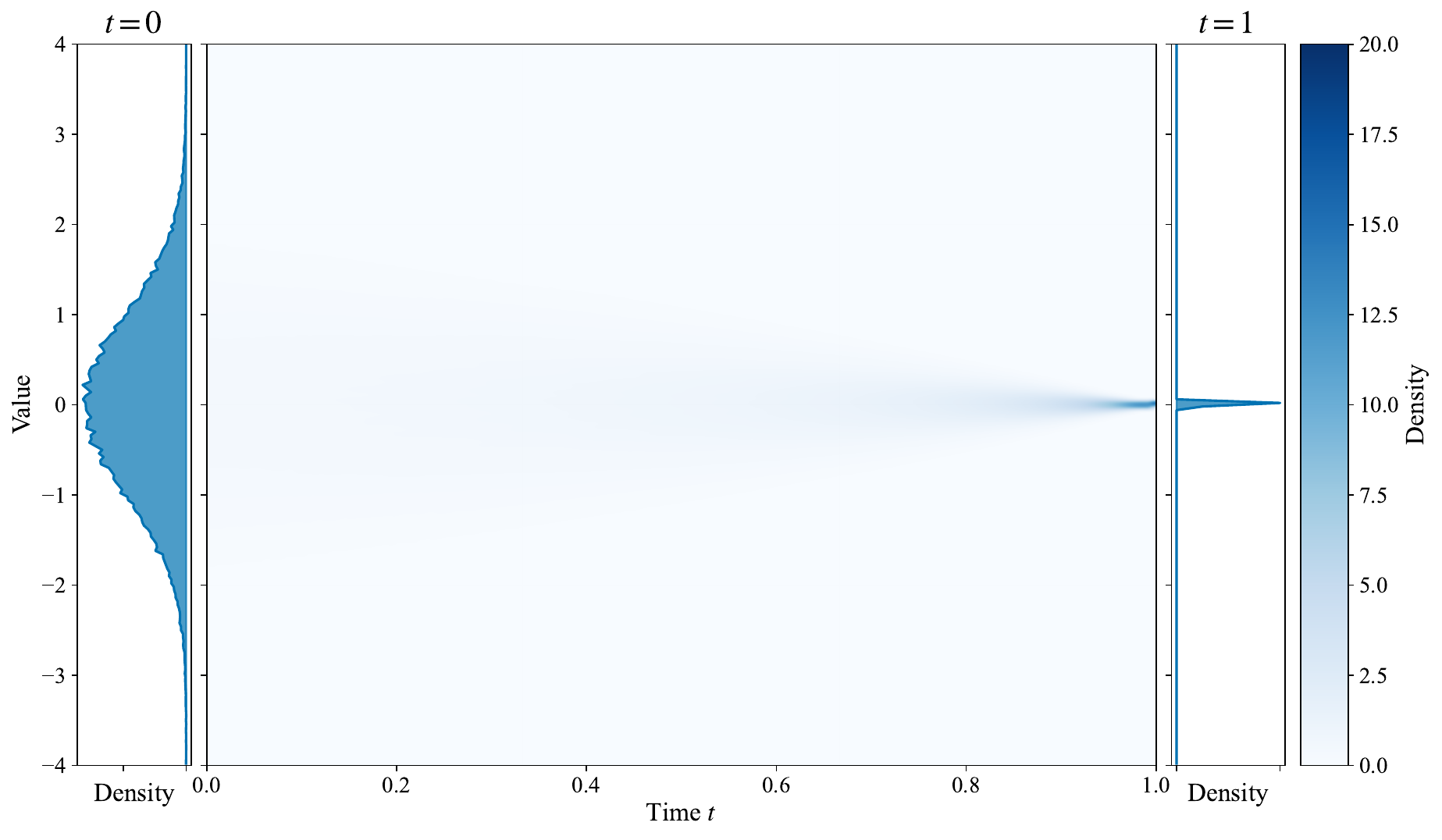}
\caption{Velocity field density demonstrating the deterministic mapping of base noise vectors into a sharp, localized delta function target.\label{fig:delta_flow_density}}
\end{figure}

Visual evidence corroborating these quantitative metrics is presented in Figures~\ref{fig:exponential_decay_result}--\ref{fig:uniform_flat_result}. Upon inspecting the generated distribution histograms, we observed highly accurate shape reconstructions across all topological presets. For successful complex cases, such as the \textsc{Triple-Mixed} scenario (Figure~\ref{fig:triple_mixed_result}), the network seamlessly learns to balance multiple overlapping Gaussian distributions with varying heights, widths, and proximity, accurately capturing their specific relative population fractions. The \textsc{Tall-Flat-Far} benchmark (Figure~\ref{fig:tall_flat_far_result}) confirms the model's ability to resolve widely separated peaks with disparate amplitudes. In the highly challenging \textsc{Noise-10Spikes} topology (Figure~\ref{fig:noise_10spikes_result}), the model precisely resolves sharp, fine-grained structural perturbations rather than artificially blurring them into a single smoothed envelope. Furthermore, Figure~\ref{fig:delta_flow_density} demonstrates the deterministic mapping of the learned velocity field in action, illustrating how the CFM vector paths intuitively transport the diffuse base Gaussian noise directly into a condensed, localized delta-function target without severe unphysical scatter. 

\begin{figure}[htbp]
\centering
\includegraphics[width=\textwidth]{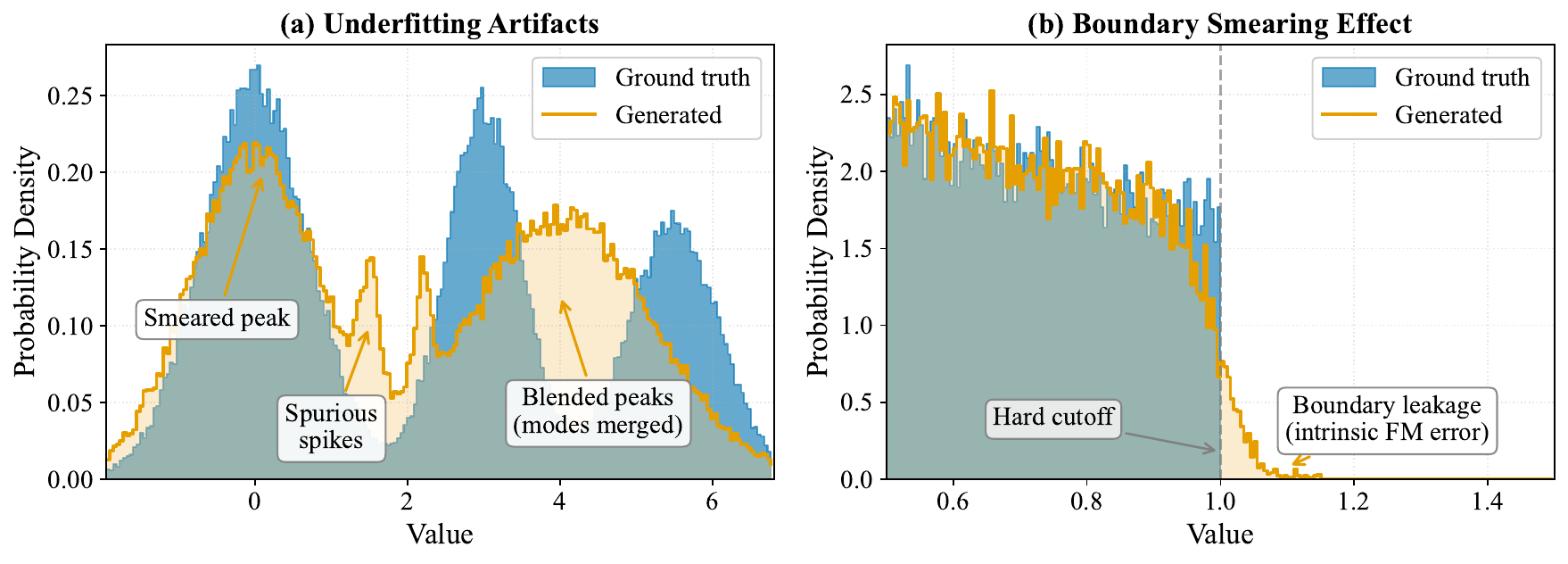}
\caption{Common generative failure modes before full convergence. (a) Underfitting artifacts exhibit smeared, spurious, and blended peaks. (b) Boundary smearing effects show slight macroscopic edge deviations along strict numerical cutoffs.\label{fig:error_types}}
\end{figure}

\begin{figure}[htbp]
\centering
\includegraphics[width=\textwidth]{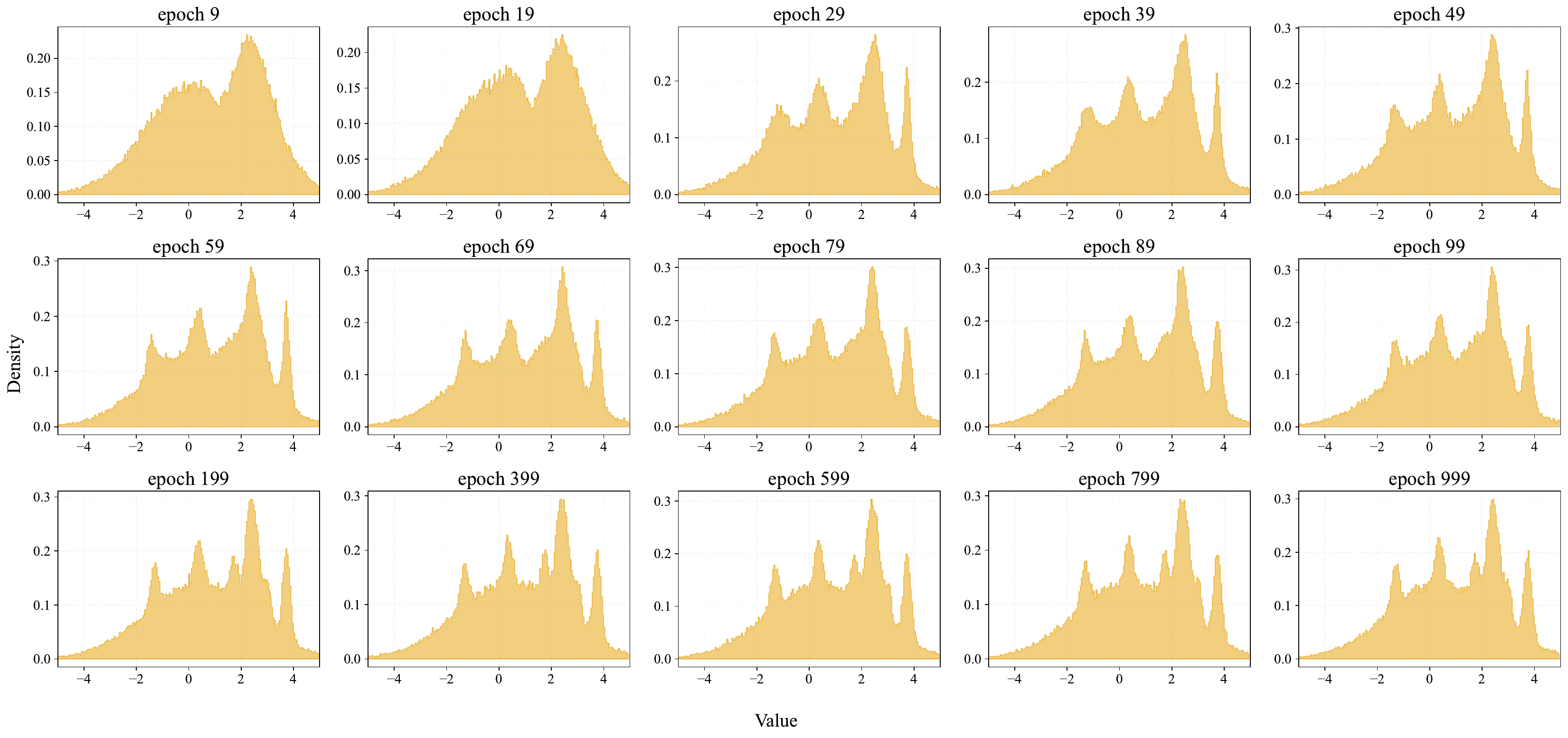}
\caption{Progressive convergence of the generated distribution evaluated on the high-frequency noise dataset. Initial smearing effects and spurious spikes systematically disappear as the model achieves convergence.\label{fig:checkpoint_evolution}}
\end{figure}

However, perfectly resolving these dense spatial features fundamentally requires decoupled validation schemas. Common physical modeling failure modes before full convergence are depicted in Figure~\ref{fig:error_types}, revealing the inherent edge-case sensitivities of flow-based generation. When undertrained, the generated manifolds frequently exhibit smeared mass boundaries along strict kinematic limits, or they produce spurious, bridged population points between completely separate disjoint peaks. Figure~\ref{fig:checkpoint_evolution} actively tracks the resolution of these artifacts by charting the progressive structural refinement over the different training stages. These sequential density profiles underscore that while the general macroscopic envelope of the distribution is identified rapidly, extended parameter optimization is strictly required for the velocity field to harden and reliably resolve targeted physical nuances.

\section{Extended unconditional generation validation}
\label{app:extended_generation}

\begin{figure}[htbp]
\centering
\includegraphics[width=\textwidth]{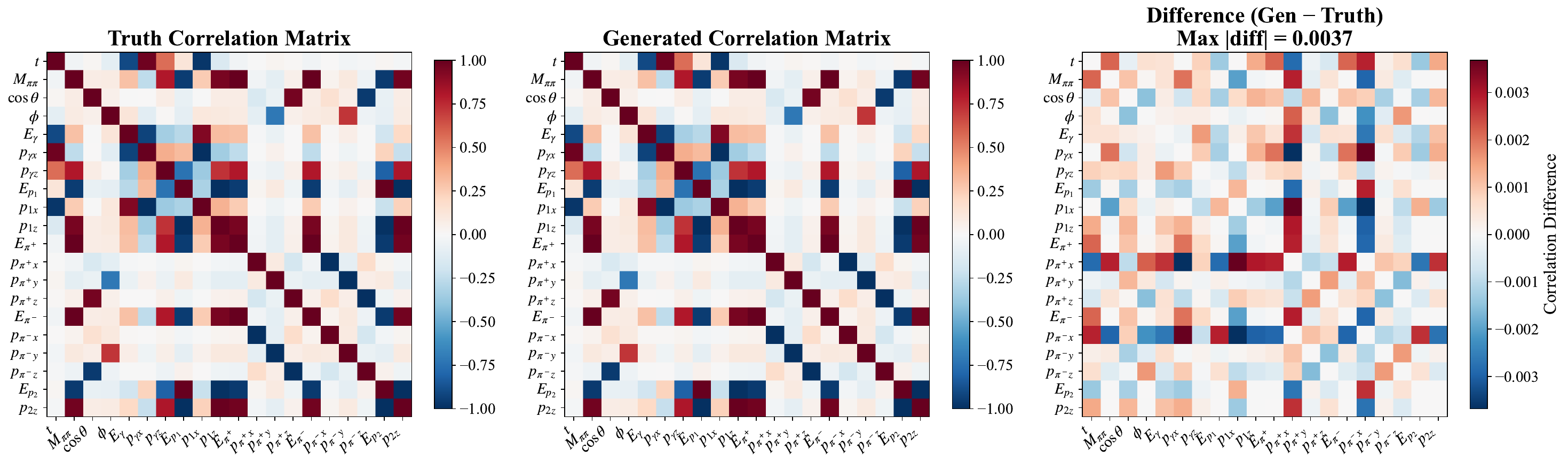}
\caption{Pearson correlation matrices for the MC-POM unconditional generation task. Assessing the dependencies between generated kinematics and the ground-truth manifold reveals excellent reproduction of complex multivariate physics constraints.\label{fig:correlation_matrix}}
\end{figure}

Beyond analyzing individual 1D marginal distributions, predicting complex multivariate correlations is crucial for validating physical simulations. Figure~\ref{fig:correlation_matrix} presents the comparative Pearson correlation matrices for both the generated and ground-truth kinematics in the unconditional MC-POM generation task. The high degree of concordance across the entire parameter space demonstrates the model's capacity to naturally reconstruct global linear relationships and couple interdependent physical limits, confirming high-fidelity multidimensional learning.

\section{Extended detector unfolding validation}
\label{app:additional_unfolding}

\begin{figure}[htbp]
\centering
\includegraphics[width=\textwidth]{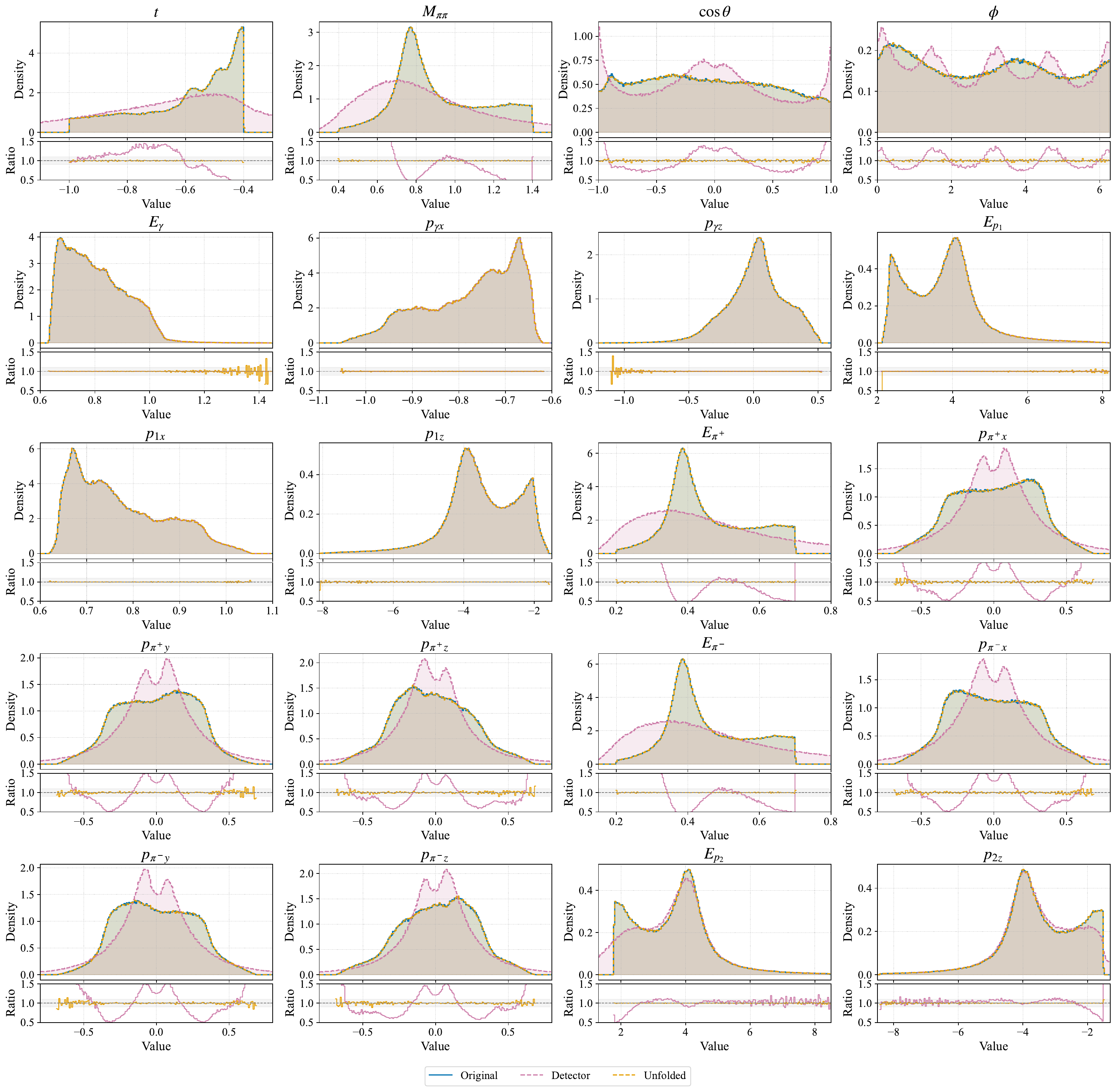}
\caption{Detector-level unfolding reconstruction initialized from severe simulated momentum smearing ($\sigma_{\mathrm{smear}}=2.0$). Despite heavy initial degradation, the generative model successfully localizes and recovers the macroscopic physical distributions.\label{fig:mcpom_denoise_sigma20}}
\end{figure}

\begin{figure}[htbp]
\centering
\includegraphics[width=\textwidth]{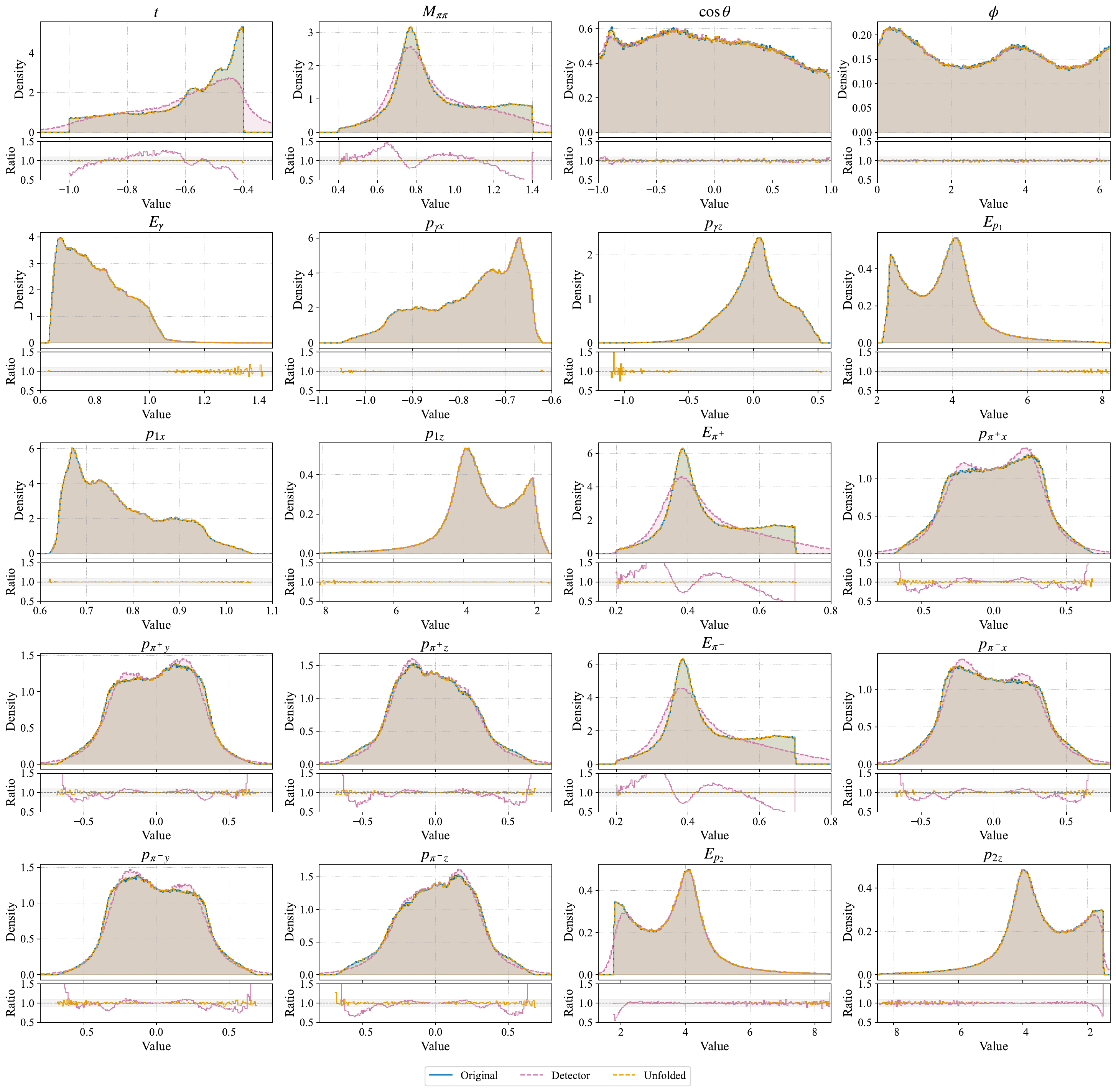}
\caption{High-fidelity detector unfolding mapped from detector uncertainties ($\sigma_{\mathrm{smear}}=0.5$). The generated recovery profiles display exceptional alignment with the particle-level parameters.\label{fig:mcpom_denoise_sigma05}}
\end{figure}

To further assess the robustness of conditional generation across varying degrees of signal degradation, we profile the unfolding consistency under diverse smearing intensities. Figure~\ref{fig:mcpom_denoise_sigma20} demonstrates the model's structural recovery when initialized with severe synthetic resolution degradation ($\sigma_{\mathrm{smear}}=2.0$), while Figure~\ref{fig:mcpom_denoise_sigma05} illustrates the near-perfect, high-fidelity phase space restoration achieved from optimally calibrated, low-uncertainty detector signals ($\sigma_{\mathrm{smear}}=0.5$). Together, these mappings highlight the stable deterministic pathways formed by the conditional CFM architecture regardless of the initial smearing scale.

\section{Computational performance}
\label{app:inference_speed}

To evaluate the computational efficiency of both training and inference, we benchmark the unconditional generation and conditional unfolding tasks on CPU and GPU hardware (full node specifications are listed in Appendix~\ref{app:hyperparameters}). Table~\ref{tab:compute_speed} summarizes the training throughput (over 10 iterations with batch size 20K) and inference throughput (over 5 repeated runs of 50K samples). The V100 GPU delivers an order-of-magnitude speedup ($\sim$32$\times$ on inference, $\sim$87$\times$ on training) over the CPU, and the A6000 further accelerates this, reaching throughputs exceeding 820K samples/s during training. These significant performance gains enable the high-throughput generation characteristic of specialized fast simulation workflows.

While training throughput is comparable across both tasks, inference speed differs dramatically: conditional unfolding achieves substantially higher throughput than unconditional generation. On the A6000, unfolding operates at $8.33 \times 10^{4}$ events/s compared to $3.04 \times 10^{3}$ events/s for unconditional generation (an $\sim$27$\times$ speed disparity consistent across GPU hardware). This disparity arises primarily from the ODE solver tolerances: unconditional generation uses strict tolerances ($\texttt{atol} = \texttt{rtol} = 10^{-7}$), whereas unfolding employs relaxed tolerances ($\texttt{atol} = \texttt{rtol} = 10^{-3}$), requiring far fewer function evaluations per integration. Notably, the relaxed tolerances introduce no measurable degradation in unfolding fidelity (Table~\ref{tab:mcpom_results}), suggesting that the conservative generation tolerances could be substantially loosened to achieve comparable speedups without compromising distributional accuracy---a promising avenue for future optimization.

\begin{table}[hbt!]
\centering
\caption{Compute speed for unconditional generation and conditional unfolding tasks across different hardware configurations.}
\label{tab:compute_speed}
\resizebox{\textwidth}{!}{%
\begin{tabular}{llcccc}
\toprule
 & & \multicolumn{2}{c}{\textbf{Training}} & \multicolumn{2}{c}{\textbf{Inference}} \\
\cmidrule(lr){3-4} \cmidrule(lr){5-6}
\textbf{Task} & \textbf{Hardware} & \textbf{ms / iter $\downarrow$} & \textbf{samples / s $\uparrow$} & \textbf{s / run $\downarrow$} & \textbf{events / s $\uparrow$} \\
\midrule
\multirow{3}{*}{Generation} & CPU (2$\times$ AMD EPYC 9454) & $2756.7 \pm 48.5$ & $7.26 \times 10^{3}$ & $635.6 \pm 1.7$ & $7.87 \times 10^{1}$ \\
 & GPU (NVIDIA Tesla V100-SXM2) & $31.7 \pm 0.9$ & $6.32 \times 10^{5}$ & $19.6 \pm 0.2$   & $2.55 \times 10^{3}$ \\
 & GPU (NVIDIA RTX A6000) & $24.3 \pm 0.1$ & $8.24 \times 10^{5}$ & $16.5 \pm 0.1$   & $3.04 \times 10^{3}$ \\
\midrule
\multirow{3}{*}{Unfolding} & CPU (2$\times$ AMD EPYC 9454) & $2907.9 \pm 21.0$ & $6.88 \times 10^{3}$ & $29.8 \pm 0.04$  & $1.68 \times 10^{3}$ \\
 & GPU (NVIDIA Tesla V100-SXM2) & $32.5 \pm 0.1$ & $6.15 \times 10^{5}$ & $0.718 \pm 0.001$ & $6.97 \times 10^{4}$ \\
 & GPU (NVIDIA RTX A6000) & $25.9 \pm 0.1$ & $7.72 \times 10^{5}$ & $0.600 \pm 0.003$ & $8.33 \times 10^{4}$ \\
\bottomrule
\end{tabular}%
}
\end{table}

\clearpage
\bibliographystyle{JHEP}
\bibliography{references}

\end{document}